\documentclass[twoside,11pt]{article}

\usepackage{jmlr2e}

\usepackage{makeidx}  
\usepackage{amsmath}
\usepackage{amsfonts}
\usepackage{amscd}
\usepackage[tableposition=top]{caption}
\usepackage{ifthen}
\usepackage[utf8]{inputenc}
\usepackage{graphicx}
\usepackage{graphicx}
\usepackage{caption}
\usepackage{subcaption}

\usepackage{color}

\usepackage{makeidx}
\usepackage{verbatim}
\usepackage{placeins}

\usepackage{amssymb}
\usepackage{bbm}

\usepackage{multirow} 
\usepackage{makecell}
\usepackage{hhline} 
\usepackage{array} 

\usepackage{siunitx}

\DeclareMathOperator*{\argmax}{arg\,max}

\definecolor{amethyst}{rgb}{0.6, 0.4, 0.8}
\definecolor{ao}{rgb}{0.0, 0.0, 1.0}

\jmlrheading{15}{2014}{0-0}{7/14}{0/00}{Lewis P. G. Evans and Niall M. Adams and Christoforos Anagnostopoulos}

\ShortHeadings{Targeting Optimal Active Learning via Example Quality}{Evans and Adams and Anagnostopoulos}
\firstpageno{1}

\begin{document}

\title{Targeting Optimal Active Learning via Example Quality}

\author{\name Lewis P. G. Evans \email lewis.evans10@imperial.ac.uk \\
       \addr Department of Mathematics\\
       Imperial College London\\
	London, SW7 2AZ, United Kingdom
       \AND
       \name Niall M. Adams \email n.adams@imperial.ac.uk \\
       \addr Department of Mathematics\\
       Imperial College London\\
	London, SW7 2AZ, United Kingdom\\
       \addr
	Heilbronn Institute for Mathematical Research\\
	University of Bristol\\
	PO Box 2495, Bristol, BS8 9AG, United Kingdom
	\AND
       \name Christoforos Anagnostopoulos \email canagnos@imperial.ac.uk \\
       \addr Department of Mathematics\\
       Imperial College London\\
	London, SW7 2AZ, United Kingdom}

\editor{Yoav Freund}

\maketitle







\begin{abstract}

In many classification problems unlabelled data is abundant and a subset can be chosen for labelling.
This defines the context of active learning (AL), where methods systematically select that subset, to improve a classifier by retraining.

Given a classification problem, and a classifier trained on a small number of labelled examples, consider the selection of a single further example.
This example will be labelled by the oracle and then used to retrain the classifier.
This example selection raises a central question: given a fully specified stochastic description of the classification problem, which example is the optimal selection?

If optimality is defined in terms of loss, this definition directly produces expected loss reduction (ELR), a central quantity whose maximum yields the optimal example selection.
This work presents a new theoretical approach to AL, \emph{example quality}, which defines optimal AL behaviour in terms of ELR.

Once optimal AL behaviour is defined mathematically, reasoning about this abstraction provides insights into AL.
In a theoretical context the optimal selection is compared to existing AL methods, showing that heuristics can make sub-optimal selections.

Algorithms are constructed to estimate example quality directly. 
A large-scale experimental study shows these algorithms to be competitive with standard AL methods.

\end{abstract}


\section{Introduction}
\label{section:Introduction}

Classification is a central task of statistical inference and machine learning.
In certain cases unlabelled data is plentiful, and a small subset can be queried for labelling.
Active learning (AL) seeks to intelligently select these unlabelled examples, to improve a base classifier.
Examples include medical image diagnosis and document categorisation \citep{Dasgupta2008,Hoi2006}.
This work focusses on the selection of a single unlabelled example, and assumes that a perfect oracle supplies the labels for selected examples.

Most AL methods are heuristic, alongside a few theoretical approaches reviewed by \citet{Settles2009,Olsson2009}.
AL method performance is often assessed by large-scale experimental studies \citep{Guyon2011,Kumar2010,Evans2013}.

A prototypical AL scenario consists of a classification problem and a classifier trained on a small labelled dataset.
The classifier may be improved by retraining on a single selected example. 
Consider the central motivating question: given a fully specified stochastic description of the classification problem, which example is the optimal selection?


Performance in classification and AL is judged by loss functions such as those described in \citet{Hand1997}, suggesting that optimality in AL selection should be defined by loss.
If optimality is defined by classifier loss, then the optimal selection is the example delivering greatest loss reduction.
Expected loss reduction (ELR) formalises the loss reduction that AL provides.
By construction, the example that maximises ELR delivers the greatest loss reduction, and is thereby the optimal AL selection in terms of classifier loss.
In this sense ELR provides a \emph{theoretical target} for AL: a theoretical quantity whose estimation will optimise performance.


This work presents a new theoretical approach to AL, example quality (EQ), where optimal AL behaviour is defined, then explored and estimated.
EQ first considers the central motivating question, then defines optimal in terms of classifier loss, since loss defines classification performance.
This definition of optimality as loss reveals ELR as the central theoretical quantity, whose maximum yields the optimal selection.
Having defined optimal AL behaviour as a mathematical abstraction, this abstraction is then explored, which generates new insights into AL.

For a classification problem with a fully specified stochastic description, EQ provides an exploration of ELR and the optimal selection.
The performance of an AL method is difficult to calculate analytically, since the selected example is usually a complicated function of the labelled data.
Even a comprehensive experimental study may fail to elucidate the performance, since there are many sources of variation including classification problem, classifier and loss.
But a full stochastic description of the problem allows examination of the AL method's selection, and comparison to the optimal selection.
This comparison is made for random selection and Shannon entropy in Section \ref{subsection:Theoretical Example to Illustrate Example Quality} (both are defined in Section \ref{subsection:Active Learning}).

EQ addresses applications by motivating the statistical estimation of ELR by new algorithms.
Algorithm development reveals surprisingly complex issues which will hopefully motivate further work to improve EQ estimation.

A large-scale experimental study evaluates the performance of EQ-estimation algorithms, alongside standard AL methods from the literature.
The study explores several sources of variation: multiple classifiers, AL algorithms, and real and theoretical classification problems (both binary and multi-class). 
The results show that the EQ-motivated algorithms perform competitively with standard AL methods.
This study finds that no single AL method is a panacea, a conclusion shared by other extensive studies of AL \citep{Guyon2011,Kumar2010,Evans2013}.

The optimal AL behaviour ELR and many AL methods condition on the labelled dataset, via their dependence on the classifier.
To address this dependence, EQ generalises ELR in expectation over the labelled data; this defines optimal AL behaviour for the classification problem itself, described in Sections \ref{subsubsection:Optimal Selection Independent of the Labelled Data} and \ref{subsubsection:The Average Performance of Shannon Entropy and Random Selection}.
This dependence also raises the new question of the robustness of the optimal AL selection to changes in the labelled data, examined in Section \ref{subsection:The Relationship between Marginal EQ and Conditional EQ}.

This work is structured as follows: first the background of classification and AL are described in Section \ref{section:Background}.
Section \ref{section:Example Quality} defines EQ, illustrated by a theoretical classification problem in Section \ref{subsection:Theoretical Example to Illustrate Example Quality}.
EQ estimation algorithms are described in Section \ref{section:Algorithms to Estimate Example Quality} and evaluated in the large-scale experimental study of Section \ref{section:Experiments and Results}, followed by concluding remarks.

\section{Background}
\label{section:Background}

The background contexts of classification and AL are described, followed by a brief review of relevant literature, with particular focus on methods that are used later in the paper.

\subsection{Classification}
\label{subsection:Classification}

A somewhat non-standard notation is developed to support this work, which stresses the dependence of the classifier on the training data.
The categorical response variable $Y$ is modelled as a function of the covariates ${\bf X}$.
For the response $Y$ there are $k$ classes with class labels $\{c_1, c_2, ..., c_k\}$. 
Each classification example is denoted $ ( {\bf x}, y) $, where ${\bf x}$ is a $d$-dimensional covariate vector and $y$ is a class label.
The class prior is denoted $\boldsymbol\pi$.

The Bayes classifier is an idealisation based on the true distributions of the classes, thereby producing optimal probability estimates, and class allocations given a loss function.
Given a covariate vector ${\bf x}$, the Bayes classifier outputs the class probability vector of $Y|{\bf x}$ denoted ${\bf p} = (p_j)_1^k$.
A \emph{probabilistic} classifier estimates the class probability vector as ${\bf \hat{p}} = (\hat{p}_j)_1^k$, and allocates ${\bf x}$ to class $\hat{y}$ using decision theoretic arguments, often using a probability threshold. 
This allocation function is denoted $h$: $\hat{y} = h({\bf \hat{p}})$.
For example, to minimise misclassification error, the most probable class is allocated: $\hat{y} = h ({\bf \hat{p}}) = \argmax_{j} (\hat{p}_j)$.
The objective of classification is to learn a rule with good generalisation properties.

A dataset is a set of examples, denoted $D = \{ {\bf x}_i,y_i \}_{i=1}^n$, where $i$ indexes the example.
This indexing notation will be useful later in the exposition.
A dataset $D_Z$ may be subdivided into training data $D_T$ and test data $D_E$.
This dataset division may be represented by index sets, for example, given an index set $Z = \{1, ..., n_z\}$ with (index) subsets $T$ and $E$, then $Z = T \cup E$ and $D_Z = D_T \cup D_E$ show the data division into training and test subsets.

First consider a parametric classifier, for example linear discriminant analysis or logistic regression \citep[Chapter~4]{Bishop2007}.
A parametric classifier has estimated parameters $\boldsymbol{\hat{\theta}}$, which can be regarded as a fixed length vector (fixed given $d$ and $k$).
These parameters are estimated by model fitting to the training data: $\boldsymbol{\hat{\theta}} = \theta (D_T)$, where $\theta$ is the model fitting function. 
This notation is intended to emphasize the dependence of the estimated parameters $\boldsymbol{\hat{\theta}}$ on the training data $D_T$.

Second, this notation is slightly abused to extend to non-parametric classifiers. 
The complexity of non-parametric classifiers may increase with sample size, hence they cannot be represented by a fixed length object.
In this case $\boldsymbol{\hat{\theta}}$ becomes a variable-length object containing the classifier's internal data (for example the nodes of a decision tree, or the stored examples of $K$-nearest-neighbours).
While the contents and meaning of $\boldsymbol{\hat{\theta}}$ would be very different, the classifier's functional roles are identical: model training produces $\boldsymbol{\hat{\theta}}$, which is used to predict class probabilities. 
These predictions are in turn used to assess classifier performance. 

To consider classifier performance, first assume a fixed training dataset $D_T$.
Classifier performance is assessed by a loss function, for example error rate, which quantifies the disagreement between the classifier's predictions and the truth. 
The empirical loss for a single example is defined via a loss function $g(y, {\bf \hat{p}})$.
Many loss functions focus on the allocated class, for example error rate $g_e(y, {\bf \hat{p}}) = \mathbbm{1} (y \ne h({\bf \hat{p}}))$.
Other loss functions focus on the predicted probability, for example log loss $g_o({\bf \hat{p}}) = \sum_{j=1}^k ( p_j \textrm{ log } \hat{p}_j )$.

The estimated probabilities ${\bf \hat{p}}$ are highly dependent on the estimated classifier $\boldsymbol{\hat{\theta}}$.
To emphasize that dependence, the empirical loss for a single example is denoted $M(\boldsymbol{\hat{\theta}}, {\bf x}, y) = g(y, {\bf \hat{p}})$.

To address the classifier's expected future loss, empirical loss is generalised to expected loss, denoted $L (\boldsymbol{\hat{\theta}})$:
\begin{equation*} 
L (\boldsymbol{\hat{\theta}}) = E_{ {\bf X},Y} [M (\boldsymbol{\hat{\theta}}, {\bf x}, y)] = E_{Y|{\bf X}} E_{\bf X} [M (\boldsymbol{\hat{\theta}}, {\bf x}, y)].
\end{equation*}
This expected loss $L$ is defined as an expectation over all possible test data.
The expected error rate and log loss are denoted $L_e$ and $L_o$.
Hereafter loss will always refer to the expected loss $L$.
The loss $L$ is dependent on the data $D$ used to train the classifier, emphasized by rewriting $L(\boldsymbol{\hat{\theta}})$ as $L(\theta(D))$ since $\boldsymbol{\hat{\theta}} = \theta(D)$.

The change in the loss as the number of labelled examples increases is of great methodological interest.
This function is known as the \emph{learning curve}, typically defined as the change of expected loss with the number of examples.
Learning curves are illustrated in Figure \ref{figure:Performance Comparison of AL and RS}, and discussed in \citet{Provost2003,Gu2001,Kadie1995}.

Section \ref{section:Experiments and Results} describes experiments with four classifiers: linear discriminant analysis, $K$-nearest-neighbours, na\"ive Bayes and support vector machine.
Linear discriminant analysis (LDA) is a linear generative classifier described in \citet[Chapter~4]{Tibshirani2009}.
$K$-Nearest-Neighbours ($K$-nn) is a well-known non-parametric classifier discussed in \citet[Chapter~4]{Duda2001}.
Na\"ive Bayes is a probabilistic classifier which assumes independence of the covariates, given the class; see \citet{Hand2001}.
The support vector machine (SVM) is a popular non-parametric classifier described in \citet{Vapnik1995}. 
Standard R implementations are used for these classifiers.

\subsection{Active Learning}
\label{subsection:Active Learning}

The context for AL is a scarcity of labelled data but an abundance of unlabelled examples. 
Good introductions to AL are provided by \citet{Dasgupta2011}, \citet{Settles2009} and \citet{Olsson2009}.

An algorithm can select a few unlabelled examples to obtain their labels from an oracle (for example a human expert).
This provides more labelled data which can be included in the training data, potentially improving a classifier.
Intuitively some examples may be more informative than others, so systematic example selection should maximise classifier improvement.

In \emph{pool-based} AL, there is an unlabelled pool of data $X_P$ from which examples may be selected for labelling. 
This pool provides a set of examples for label querying, and also gives further information on the distribution of the covariates.
Usually there is also a (relatively small) dataset of labelled examples, denoted $D_S$, of size $n_s$.
This work considers the scenario of pool-based AL.

In AL it is common to examine the learning curve, by repeating the AL selection step many times (\emph{iterated AL}).
At each selection step, the loss is recorded, and this generates a set of losses, which define the learning curve for the AL method.
Iterated AL allows the exploration of performance over the learning curve, as the amount of labelled data grows.
This repeated application of AL selection is common in both applications and experimental studies \citep{Guyon2011,Evans2013}.
A recurring theme in AL is the strong dependence of the classifier on the training data, and iterated AL provides a detailed examination of this dependence.

At each selection step, an AL method may select a single example from the pool (\emph{sequential AL}) or several examples at once (\emph{batch AL}).
AL applications are often constrained to use batch AL for pragmatic reasons \citep{Settles2009}.

Turning to AL performance, consider random selection (RS) where examples are chosen randomly (uniformly) from the pool.
By contrast, AL methods select some examples in preference to others.
Under RS and AL, the classifier receives exactly the same number of labelled examples; thus RS provides a reasonable benchmark for AL \citep{Guyon2011,Evans2013}.
The comparison of methods to benchmarks is available in experiments but not in AL applications \citep{Provost2010}.

\begin{figure}
\centering
\includegraphics[scale=0.8]{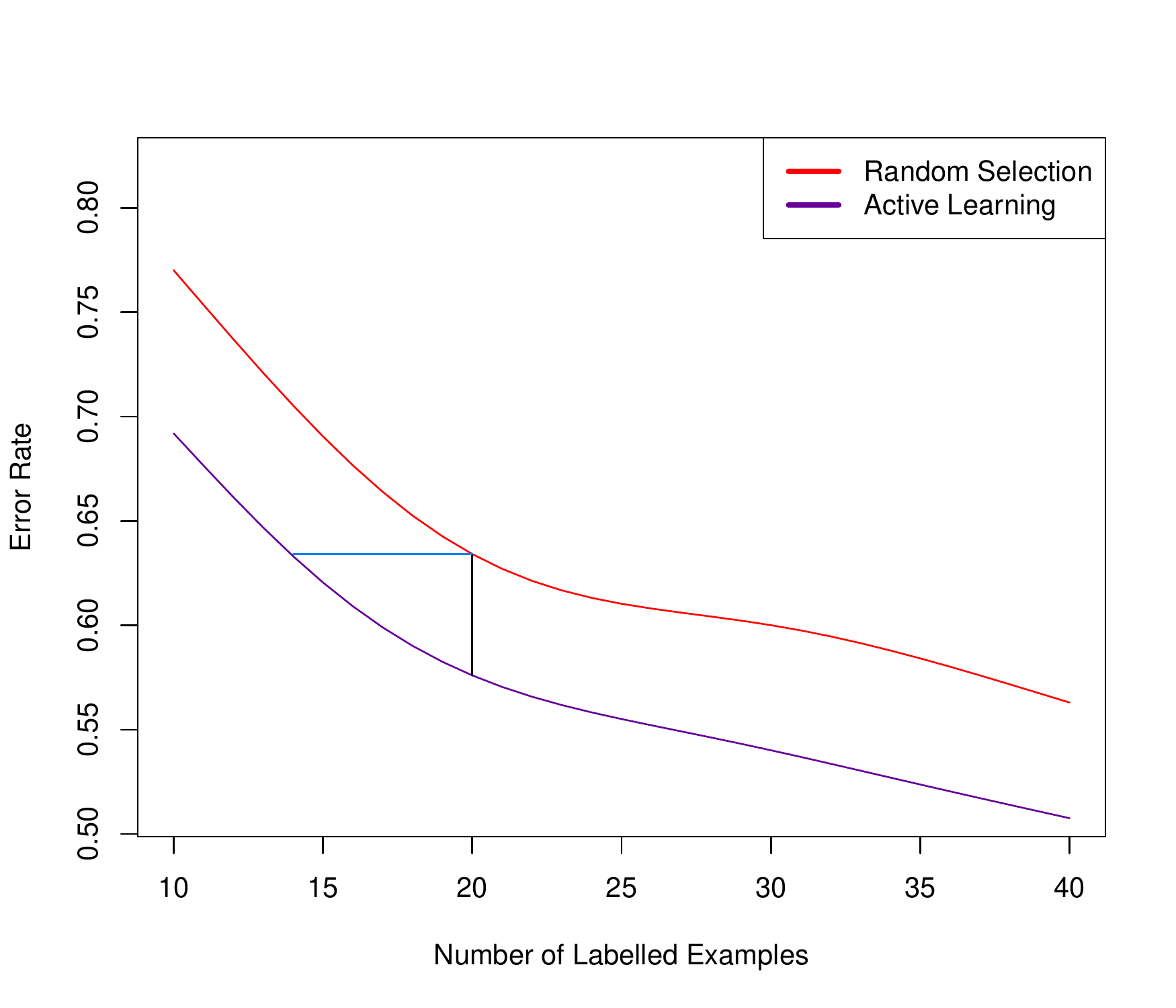}
\caption{Performance comparison of active learning and random selection, showing that a classifier often improves faster under AL than under RS.
In both cases the loss decreases as the number of labelled examples increases; however, AL improves faster than RS. 
These curves are smoothed averages from multiple experiments. 
The black vertical line illustrates the fixed-label comparison, whereas the blue horizontal line shows the fixed-loss comparison (see Section \ref{subsection:Active Learning}). 
The classification problem is ``Abalone'' from UCI, a three-class problem, with the base classifier being $5$-nn, and the AL method being Shannon entropy.
}
\label{figure:Performance Comparison of AL and RS}
\end{figure}

Classifier performance (usually) improves even under the benchmark RS, since the classifier receives more training data.
AL performance assessment should consider how much AL outperforms RS. 
Hence AL performance addresses the relative improvement of AL over RS, and the relative ranks of AL methods, rather than the absolute level of classifier performance.
Figure \ref{figure:Performance Comparison of AL and RS} shows the losses of AL and RS as the number of labelled examples increases.

Figure \ref{figure:Performance Comparison of AL and RS} shows two different senses in which AL outperforms RS: first AL achieves better loss reduction for the same number of labels (\emph{fixed-label comparison}), and second AL needs fewer labels to reach the same classifier performance (\emph{fixed-loss comparison}).
Together the fixed-label comparison and fixed-loss comparison form the two fundamental aspects of AL performance.
The fixed-label comparison first fixes the number of labels, then seeks to minimise loss.
Several established performance metrics focus on the fixed-label comparison: AUA, ALC and WI \citep{Guyon2011,Evans2013}.
The fixed-label comparison is more common in applications where the costs of labelling are significant \citep{Settles2009}.

Under the fixed-loss comparison, the desired level of classifier loss is fixed, the goal being to minimise the number of labels needed to reach that level.
Label complexity is the classic example, where the desired loss level is a fixed ratio of asymptotic classifier performance \citep{Dasgupta2011}.
Label complexity is often used as a performance metric in contexts where certain assumptions permit analytically tractable results, for example \citet{Dasgupta2011}.

\subsection{Overview of Active Learning Methods}
\label{subsection:Literature Review}

Uncertainty sampling is a heuristic approach to AL, where examples are chosen closest to the classifier's estimated decision boundary \citep{Thrun1992,Settles2009}.
This approach selects examples of the greatest classifier uncertainty in terms of class membership probability.
The idea is that these uncertain examples will be the most useful for tuning the classifier's decision boundary.
Example methods include Shannon entropy (SE), least confidence and maximum uncertainty.
For a single unlabelled example {\bf x}, least confidence is defined as
$ U_L ({\bf x}, \theta(D)) = 1 - \hat{p}(\hat{y} | {\bf x})$,
where $\hat{p}(\hat{y} | {\bf x})$ is the classifier's estimated probability of the allocated class $\hat{y}$.
Shannon entropy is defined as
$ U_E ({\bf x}, \theta(D)) = \sum_{j=1}^k \hat{p}_j \, \textrm{log} (\hat{p}_j)$.
The uncertainty sampling approach is popular and efficient, but lacks theoretical justification.

Version space search is a theoretical approach to AL, where the version space is the set of hypotheses (classifiers) that are consistent with the data \citep{Mitchell1997,Dasgupta2011}.
Learning is then interpreted as a search through version space for the optimal hypothesis.
The central idea is that AL can search this version space more efficiently than RS.

Query by committee (QBC) is a loose approximation to version space search \citep{Seung1992}.
Here a committee of classifiers is trained on the labelled data, which then selects the unlabelled examples where the committee's predictions disagree the most.
This prediction disagreement may focus on predicted classes (for example vote entropy) or predicted class probabilities (for example average Kullback-Liebler divergence); see \citet{Olsson2009}.
These widely used versions of QBC are denoted QbcV and QbcA.
A critical control parameter for QBC is the choice of classifier committee, which lacks theoretical guidance.
In this sense version space search leaves the optimal AL selection unspecified.

Another approach to AL is exploitation of cluster structure in the pool.
Elucidating the cluster structure of the pool could provide valuable insights for example selection.
\citet{Dasgupta2011} gives a motivating example: if the pool clusters neatly into $b$ class-pure clusters where $b=k$, then $b$ labels could suffice to build an optimal classifier.
This very optimistic example does illustrate the potential gain.

A third theoretical approach is error reduction, introduced in \citet{Roy2001}.
This approach minimises the error of the retrained classifier, which is the error of the classifier which has been retrained on the selected example.
Those authors consider two loss functions, error rate and log loss, to construct two quantities, which are referred to here as expected future error (EFE) and expected future log loss (EFLL).
Those authors focus on methods to estimate EFE and EFLL, before examining the experimental performance of their estimators.

The error reduction approach is similar in spirit to example quality, since the optimal example selection is first considered, and then specified in terms of classifier loss.
The quantity EFE is a valuable precursor to expected loss reduction, being a component of ELR, which is defined in Equation \ref{eq:eqc}.
However EFE omits the loss of the current classifier, which proves important when taking expectation over the labelled data (see Sections \ref{section:Example Quality} and \ref{subsection:The Relationship between Marginal EQ and Conditional EQ}) and when examining improvement (see Section
\ref{subsection:Theoretical Example to Illustrate Example Quality}).
Those authors focus on EFE estimation, rather than using EFE and EFLL to construct a theoretical account of AL.


Given a classifier fitting function $\theta$, labelled data $D$ and a single unlabelled example ${\bf x}$, EFE is defined as 
\begin{equation*} 
EFE({\bf x}, \theta, D) = - E_{Y | {\bf x}} [ L_e(\theta(D \cup ({\bf x},Y)) ] = - \sum_{j=1}^k \{ p_j \, L_e(\theta(D \cup ({\bf x},c_j)) \},
\end{equation*}
where $L_e$ is error rate (see Section \ref{subsection:Classification}).
EFLL is defined in the very same way as EFE, with log loss $L_o$ replacing error rate $L_e$.
Both of these quantities average over the unobserved label $Y|{\bf x}$.

Those authors define an algorithm to calculate EFE, denoted EfeLc, which approximates the loss using the unlabelled pool for efficiency.
Specifically it approximates error rate $L_e$ by the total least confidence over the entire pool:
\begin{equation*} 
L_e(\theta(D)) \approx \sum_{ {\bf x}_i \in X_P} U_L({\bf x}_i, \theta(D)),
\end{equation*}
where $X_P$ are the unlabelled examples in the pool.

Those authors propose the following approximation for the value of EFE by calculating
\begin{equation} \label{eq:efelc}
\begin{split}
f_1({\bf x}, \theta, D) = - \sum_{j=1}^k \left\{ \hat{p}_j \sum_{ {\bf x}_i \in X_P} U_L({\bf x}_i, \theta(D \cup ({\bf x}_i,c_j))) \right\} 
= - \sum_{j=1}^k \left\{ \hat{p}_j \sum_{ {\bf x}_i \in X_P} \left( 1 - \hat{p}(\hat{y_i} | {\bf x}_i) \right) \right\}.
\end{split}
\end{equation}

This approximation of $L_e$ by the total least confidence over the pool is potentially problematic.
It is easy to construct cases (for example an extreme outlier) where a labelled example would reduce a classifier's uncertainty, but also increase the overall error; such examples call into question the approximation of error by uncertainty.
In the absence of further assumptions or motivation, it is hard to anticipate the statistical properties of $f_1$ in Equation \ref{eq:efelc} as an estimator.

\section{Example Quality}
\label{section:Example Quality}

Here the theoretical target, example quality, is defined as expected loss reduction.
This motivates EQ as an estimation target, both theoretically and for applications.

\subsection{The Definition of Example Quality}
\label{subsection:The Definition of Example Quality}

To define the theoretical target, knowledge is assumed of the underlying distribution $({\bf X}, Y)$, with all expectations being formed with respect to that joint distribution.
Assume a fixed dataset $D_S$ sampled i.i.d. from the joint distribution $({\bf X}, Y)$.
The dependence of the classifier $\boldsymbol{\hat{\theta}}$ on the data $D_S$ is critical, with the notation $\boldsymbol{\hat{\theta}} = \theta(D_S)$ intended to emphasize this dependence.

First assume a base classifier already trained on a dataset $D_S$.
Consider how much a single labelled example improves performance.
The single labelled example $({\bf x},y)$ will be chosen from a \emph{labelled} dataset $D_W$, to develop the argument.
This loss from retraining on that single labelled example is examined in order to later define the loss for the expected label of an unlabelled example.

Consider the selection of a single labelled example $({\bf x},y)$ from $D_W$, given the labelled data $D_S$, the classifier training function $\theta$ and a loss function $L$.	
The reduction of the loss for retraining on that example is defined as actual-EQ, denoted $Q^a$:

\begin{equation*} 
Q^a({\bf x}, y, \theta, D_S) = L(\theta(D_S)) - L(\theta(D_S \cup ({\bf x},y)).
\end{equation*}

$Q^a$ is the actual classifier improvement from retraining on the labelled example $({\bf x},y)$.
The goal here is to maximise the reduction of loss.
The greatest loss reduction is achieved by selecting the example $({\bf x_*},y_*)$ from $D_W$ that maximises $Q^a$, given by
\begin{equation*} 
{(\bf x_*}, y_*) = \argmax_{ ({\bf x}, y) \in D_W} Q^a({\bf x}, y, \theta, D_S).
\end{equation*} 


Turning to AL, the single example ${\bf x}$ is unlabelled, and will be chosen from the unlabelled pool $X_P$.
Knowledge of the underlying joint distribution $({\bf X}, Y)$ is still assumed.
Here the unknown label of ${\bf x}$ is a random variable, $Y|{\bf x}$, and taking its expectation allows the expected loss to defined, this being the classifier loss after retraining with the unlabelled example and its unknown label.
Thus the expected loss is defined using the expectation over the label $Y|{\bf x}$ to form conditional-EQ, denoted $Q^c$:

\begin{equation} \label{eq:eqc}
\begin{split}
Q^c({\bf x}, \theta, D_S) = E_{Y | {\bf x}} [Q^a({\bf x}, Y, \theta, D_S)] = L(\theta(D_S)) - E_{Y | {\bf x}} [ L(\theta(D_S \cup ({\bf x},Y)) ] \\
= \underbrace{ L(\theta(D_S)) }_{\text{Term F}} - \underbrace{ \sum_{j=1}^k \{ \underbrace{ p_j \, L(\theta(D_S \cup ({\bf x},c_j)) }_{\text{Term H}} \} }_{\text{Term J}}.
\end{split}
\end{equation}

$Q^c$ defines the expected loss reduction from retraining on the unlabelled example ${\bf x}$ with its unknown label $Y|{\bf x}$.
In other words $Q^c$ captures the difference between two losses, of the existing classifier against the expected loss of the retrained classifier.
In this sense $Q^c$ is the improvement function, since it defines exactly how much this example will improve the classifier.


The unlabelled example ${\bf x_*}$ from the pool $X_P$ that maximises $Q^c$ is the optimal example selection:
\begin{equation*} 
{\bf x_*} = \argmax_{ {\bf x} \in X_P} Q^c({\bf x}, \theta, D_S).
\end{equation*}

Algorithms to estimate the target $Q^c$ are presented in Section \ref{section:Algorithms to Estimate Example Quality}.

The target $Q^c$ extends to define the optimal selection for batch AL, but this extension is omitted for space.




For a theoretical classification problem, the target $Q^c$ can be evaluated exactly, to reveal the best and worst possible loss reduction (in expectation), by maximising and minimising $Q^c$.
Figure \ref{figure:Maximum and minimum AL performance with simulated data, exact EQ} shows that the best and worst AL performance curves are obtained by maximising and minimising $Q^c$.

Given a theoretical classification problem, the target $Q^c$ also provides straightforward calculations for the performance of random selection, and the regret of AL methods such as Shannon entropy; these calculations are omitted for space.

\begin{figure}
\centering
\includegraphics[scale=0.8]{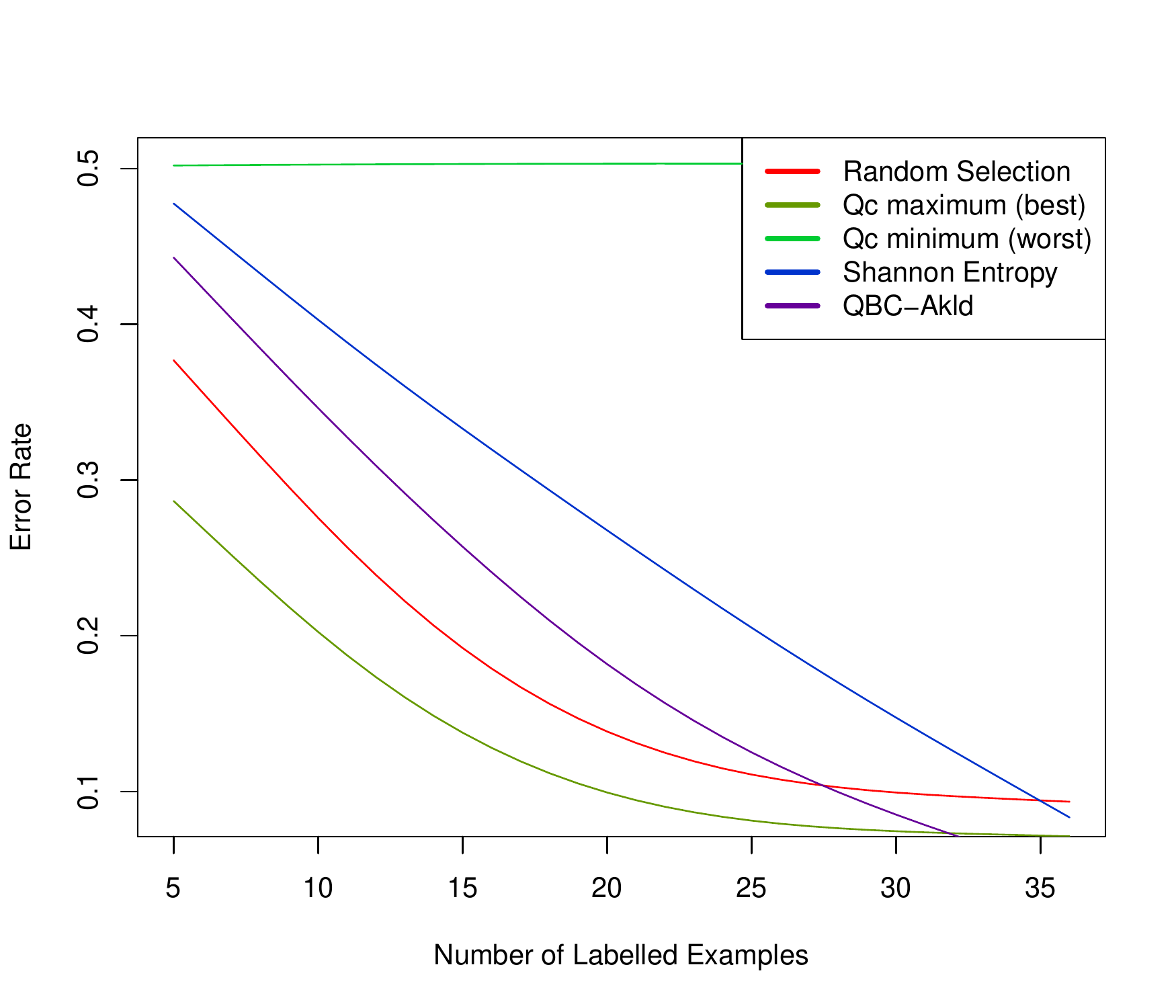}
\caption{The best and worst AL performance curves are obtained by maximising and minimising the target $Q^c$, which demonstrate the extremes of AL performance.
With simulated data, $Q^c$ can be calculated exactly; here the classification problem is the Ripley four-Gaussian problem (illustrated in Figure \ref{fig:Qc rankings of Ripley to show similarity of rankings 1___Truth}).
These curves are smoothed from multiple experiments, with the base classifier being $5$-nn.
}
\label{figure:Maximum and minimum AL performance with simulated data, exact EQ}
\end{figure}


\subsubsection{Optimal Selection Independent of the Labelled Data}
\label{subsubsection:Optimal Selection Independent of the Labelled Data}

Here the optimal AL behaviour is examined, independently of the labelled data $D_S$.
This provides further insight into the optimal AL selection for the classification problem itself.

The target $Q^c$ conditions on the data $D_S$, which is a primary source of variation for $Q^c$.
To address this source of variation, the expectation over $D_S$ is taken, leading to marginal-EQ, denoted $Q^m$:
\begin{equation} \label{eq:eqm}
Q^m({\bf x}, \theta, n_s) = E_D [ Q^c({\bf x}, \theta, D) ] = E_D [ L(\theta(D)) - E_{Y | {\bf x}} [ L(\theta(D \cup ({\bf x},Y)) ] ],
\end{equation}
where $n_s$ is the constant size of the dataset $D_S$: $n_s = |D_S|$.

$Q^m$ is the marginal improvement expected for a classifier retrained on the unlabelled example ${\bf x}$, in expectation over a dataset $D_S$ drawn i.i.d. from $({\bf X}, Y)$. 
The dataset size $n_s$ still matters: as $n_s\to \infty$, the ranking signal of $Q^m$ vanishes, as all examples provide equal (zero) loss reduction.

The unlabelled example that maximises $Q^m$  is denoted ${\bf x_+}$:
\begin{equation*}
{\bf x_+} = \argmax_{ {\bf x} \in X_P} Q^m({\bf x}, \theta, n_s).
\end{equation*}

This marginal target $Q^m$ defines optimal AL behaviour for the underlying classification problem itself, independently of $D_S$.
The targets $Q^c$ and $Q^m$, with their maxima $x_*$ and $x_+$, reveal optimal AL behaviour, whether conditioning on a single dataset ($Q^c$) or whether considering the classification problem as a whole ($Q^m$).

$Q^m$ is calculated and illustrated for a theoretical classification problem in Section \ref{subsubsection:The Average Performance of Shannon Entropy and Random Selection}.
The target $Q^m$ raises another question for AL: exactly how $Q^c$ depends on $D_S$, and the consequences for optimal AL behaviour; this is addressed in Section \ref{subsection:The Relationship between Marginal EQ and Conditional EQ}.


The central motivating question from The Introduction, which example is the optimal selection, can now be generalised to form a second central question: which example is the optimal selection, for the classification problem itself, independently of $D_S$?
This second question is answered by the marginal target $Q^m$ and its maximum $x_+$, illustrated in Figures \ref{fig:Se-and-Rs-and-Qm for two Examples}, \ref{fig:Qc rankings of Ripley to show similarity of rankings 1___averagedRanks} and \ref{fig:Qm for two loss functions}.
This second question creates a new shift in AL from conditional-AL ($Q^c$, $x_*$) to marginal AL ($Q^m$, $x_+$).

\subsection{Theoretical Example}
\label{subsection:Theoretical Example to Illustrate Example Quality}

An example using a theoretical classification problem is presented, to illustrate EQ in detail.
The stochastic character of this problem is fully specified, allowing exact calculations of the loss $L$, and the theoretical targets $Q^c$ and $Q^m$ as functions of the univariate covariate $x$.
To reason about $Q^c$ as a function of $x$, an infinite pool is assumed, allowing any $x \in \mathbb{R}$ to be selected.
These targets are then explored as functions of $x$, and the optimal AL selection $x_*$ is examined, where $x_*= \argmax_x {Q^c(x)}, x \in \mathbb{R}$.

When the full stochastic description of the problem is known, this knowledge allows examination of the AL method's selection, denoted $x_r$, and comparison to the optimal selection $x_*$.
This comparison is made below for the popular AL heuristic Shannon entropy, and for random selection.

Consider a binary univariate problem, defined by a balanced mixture of two Gaussians: $ \{ {\boldsymbol\pi} = (\frac{1}{2}, \frac{1}{2}), (X|Y=c_1) \sim \textrm{N}(-1,1), (X|Y=c_2) \sim \textrm{N}(1,1) \}$.
The prior ${\boldsymbol\pi}$ is fixed here (though varied later to illustrate the target $Q^m$, see Section \ref{subsubsection:The Average Performance of Shannon Entropy and Random Selection}).
The loss function is error rate $L_e$ (defined in Section \ref{subsection:Classification}).
The true decision boundary to minimise error rate is denoted $t = \frac{1}{2} ({\mu}_1 + {\mu}_2)$.

Every dataset $D$ of size $n$ sampled from this problem is assumed to split equally into two class-pure subsets $D_j = \{ y_i = c_j, (x_i, y_i) \in D \}$ each of size $n_j = \frac{n}{2}$; this is sampling while holding the prior fixed.
Appendix A provides full calculation details for this example.

Consider a classifier that estimates only the class-conditional means, given the true prior $\boldsymbol\pi$ and the true common variance of 1.
The classifier parameter vector is $\boldsymbol{\hat{\theta}} = ( \hat{\mu}_1, \hat{\mu}_2 )$, where $\hat{\mu}_j$ is the sample mean for class $c_j$.
The classifier's estimated decision boundary to minimise error rate is denoted $\hat{t} = \frac{1}{2} (\hat{\mu}_1 + \hat{\mu}_2)$.


\subsubsection{Calculation and Exploration of $Q^c$}
\label{subsubsection:Calculation of Qc}

$Q^c$ is calculated exactly as a function of $x$, to explore $Q^c$ as a function of $x$, and to examine the optimal selection $x_*$.

The classifier's decision rule $r_1(x)$ minimises the loss $L_e(\boldsymbol{\hat{\theta}})$, and is given in terms of a threshold on the estimated class probabilities by
\begin{displaymath}
   r_1(x) = \left\{
     \begin{array}{lr}
       \hat{y} = c_1 & : \hat{p}_1(x) > \frac{1}{2}, \\
       \hat{y} = c_2 & : \hat{p}_1(x) < \frac{1}{2}, \\
     \end{array}
   \right.
\end{displaymath}
or equivalently, in terms of a decision boundary on $x$, by
\begin{displaymath}
   r_1(x) = \left\{
     \begin{array}{lr}
	\hat{\mu}_1 < \hat{\mu}_2 & : \hat{y} = c_1 \textrm{ if } x < \hat{t}, c_2 \textrm{ otherwise}, \\
	\hat{\mu}_1 > \hat{\mu}_2 & : \hat{y} = c_1 \textrm{ if } x > \hat{t}, c_2 \textrm{ otherwise}. \\
     \end{array}
   \right.
\end{displaymath}


The classifier may get the estimated class means the wrong way around, in the unlikely case that $\hat{\mu}_1 > \hat{\mu}_2$.
As a result the classifier's behaviour is very sensitive to the condition $(\hat{\mu}_1 > \hat{\mu}_2)$, as shown by the second form of the decision rule $r_1(x)$, and by the loss function in Equation \ref{eq:le_equation_1}.

It is straightforward to show that the loss $L_e(\boldsymbol{\hat{\theta}})$ is given by
\begin{equation} \label{eq:le_equation_1}
\begin{split}
L_e(\boldsymbol{\hat{\theta}}) = \frac{1}{2} \{ 1 - F_1(\hat{t}) + F_2(\hat{t}) + \mathbbm{1}(\hat{\mu}_1 > \hat{\mu}_2) [2 F_1(\hat{t}) - 2 F_2(\hat{t})] \},
\end{split}
\end{equation}
where $F_j(x)$ denotes the cdf for class-conditional $(X|Y=c_j)$.
This result is derived in Appendix A.

In AL an unlabelled point $x$ is chosen for the oracle to label, before retraining the classifier.
Retraining the classifier with a single new example $(x, c_j)$ yields a new parameter estimate denoted $\boldsymbol{\hat{\theta}}^{\prime}_j$, where the mean estimate for class $c_j$ has a new value denoted $\hat{\mu}^{\prime}_j$, with a new estimated boundary denoted $\hat{t}^{\prime}_j$.

Here $\hat{\mu}^{\prime}_j = (1-z)\hat{\mu}_j + z x$ where $z = \frac{2}{n+2}$, $z$ being an updating constant which reflects the impact of the new example on the mean estimate $\hat{\mu}_j$.

To calculate $Q^c$ under error loss $L_e$, observe that the Term J from Equation \ref{eq:eqc} is $[ p_1 L_e(\boldsymbol{\hat{\theta}}^{\prime}_1) + p_2 L_e(\boldsymbol{\hat{\theta}}^{\prime}_2) ]$.
Term F in Equation \ref{eq:eqc} is directly given by Equation \ref{eq:le_equation_1}.
From Equations \ref{eq:eqc} and \ref{eq:le_equation_1}, $Q^c (x, \theta, D) = Q^c(x, \boldsymbol{\hat{\theta}})= L_e(\boldsymbol{\hat{\theta}}) - [ p_1 L_e(\boldsymbol{\hat{\theta}}^{\prime}_1) + p_2 L_e(\boldsymbol{\hat{\theta}}^{\prime}_2) ]$, hence
\begin{equation*} 
\begin{split}
Q^c (x, \boldsymbol{\hat{\theta}}) = Q^c (x, (\hat{\mu}_1, \hat{\mu}_2)) = \frac{1}{2} \{ 1 - F_1(\hat{t}) + F_2(\hat{t}) + \mathbbm{1}(\hat{\mu}_1 > \hat{\mu}_2) [2 F_1(\hat{t}) - 2 F_2(\hat{t})] \} \\
 - \frac{p_1}{2} \{ 1 - F_1(\hat{t}^{\prime}_1) + F_2(\hat{t}^{\prime}_1) + \mathbbm{1}(\hat{\mu}^{\prime}_1 > \hat{\mu}_2) [2 F_1(\hat{t}^{\prime}_1) - 2 F_2(\hat{t}^{\prime}_1)] \} \\
 - \frac{p_2}{2} \{ 1 - F_1(\hat{t}^{\prime}_2) + F_2(\hat{t}^{\prime}_2) + \mathbbm{1}(\hat{\mu}_1 > \hat{\mu}^{\prime}_2) [2 F_1(\hat{t}^{\prime}_2) - 2 F_2(\hat{t}^{\prime}_2)] \},
\end{split}
\end{equation*}
where $p_j$, $\hat{\mu}^{\prime}_j$, and $\hat{t}^{\prime}_j$ are functions of $x$.

Even for this simple univariate problem, $Q^c(x, \boldsymbol{\hat{\theta}})$ is a complicated non-linear function of $x$.
Given the difficulty of analytically analysing and optimising $Q^c$, specific cases of $\boldsymbol{\hat{\theta}}$ allow exploration of $Q^c$, shown in Figure \ref{fig:Qc for specific thetaHat cases, Example 1}.
In each specific case of $\boldsymbol{\hat{\theta}}$, $x_*$ yields greatest correction to $\boldsymbol{\hat{\theta}}$ in terms of moving the estimated boundary $\hat{t}$ closer to the true boundary $t$. 
This is intuitively reasonable since error rate is a function of $\hat{t}$ and minimised for $\hat{t} = t$.

In the first two cases (Figures \ref{fig:Qc for specific thetaHat cases, Example 1_case_mu1h=[-0_5]__mu2h=[1_5]} and \ref{fig:Qc for specific thetaHat cases, Example 1_case_mu1h=[-0_9]__mu2h=[1_1]}), $x_*$ is negative, to improve the classifier by reducing the overestimate of $\hat{t}$.
In the third case (Figure \ref{fig:Qc for specific thetaHat cases, Example 1_case_mu1h=[-1_1]__mu2h=[1_1]}), $\hat{t} = t$ and here the classifier's loss $L_e$ cannot be reduced, shown by $Q^c(x) < 0$ for all $x$. 
The fourth case (Figure \ref{fig:Qc for specific thetaHat cases, Example 1_case_mu1h=[1]__mu2h=[-1]}) is interesting because the signs of the estimated means are reversed compared to the true means, and here the most non-central $x$ offer greatest classifier improvement.
Together these cases show that even for this toy example, the improvement function $Q^c$ is complicated and highly dependent on the estimated parameters.

\begin{figure}
        \centering
        \begin{subfigure}[b]{0.45\textwidth}
                \centering
               \includegraphics[width=\textwidth]{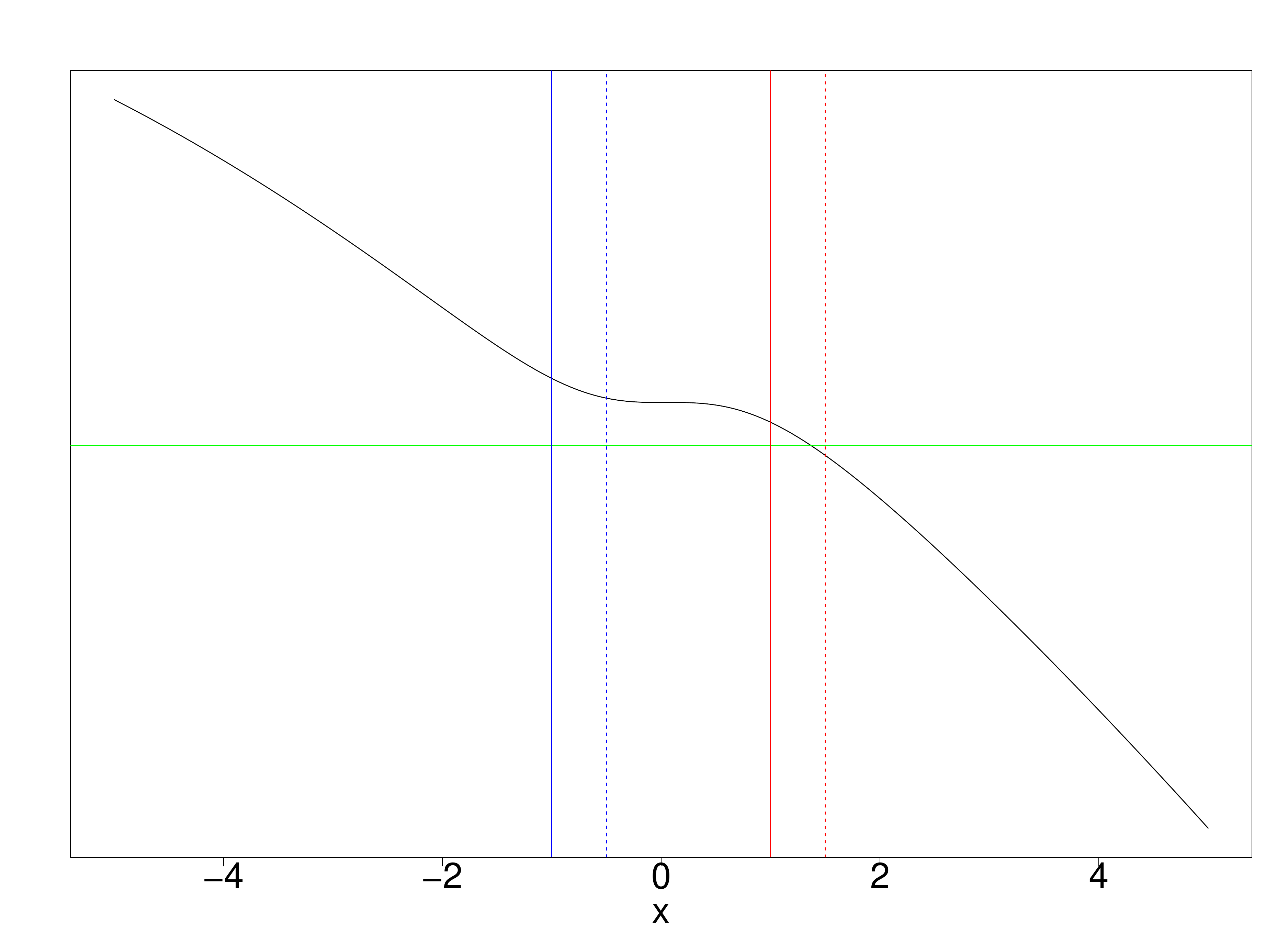}
                \caption{$\boldsymbol{\hat{\theta}} = (\hat{\mu}_1=-0.5,\hat{\mu}_2=1.5)$; \\
$\hat{\mu}_1$, $\hat{\mu}_2$ are right-shifted, $\hat{\mu}_j = {\mu}_j + 0.5$ \\}
                \label{fig:Qc for specific thetaHat cases, Example 1_case_mu1h=[-0_5]__mu2h=[1_5]}
        \end{subfigure}
        \begin{subfigure}[b]{0.45\textwidth}
                \centering
               \includegraphics[width=\textwidth]{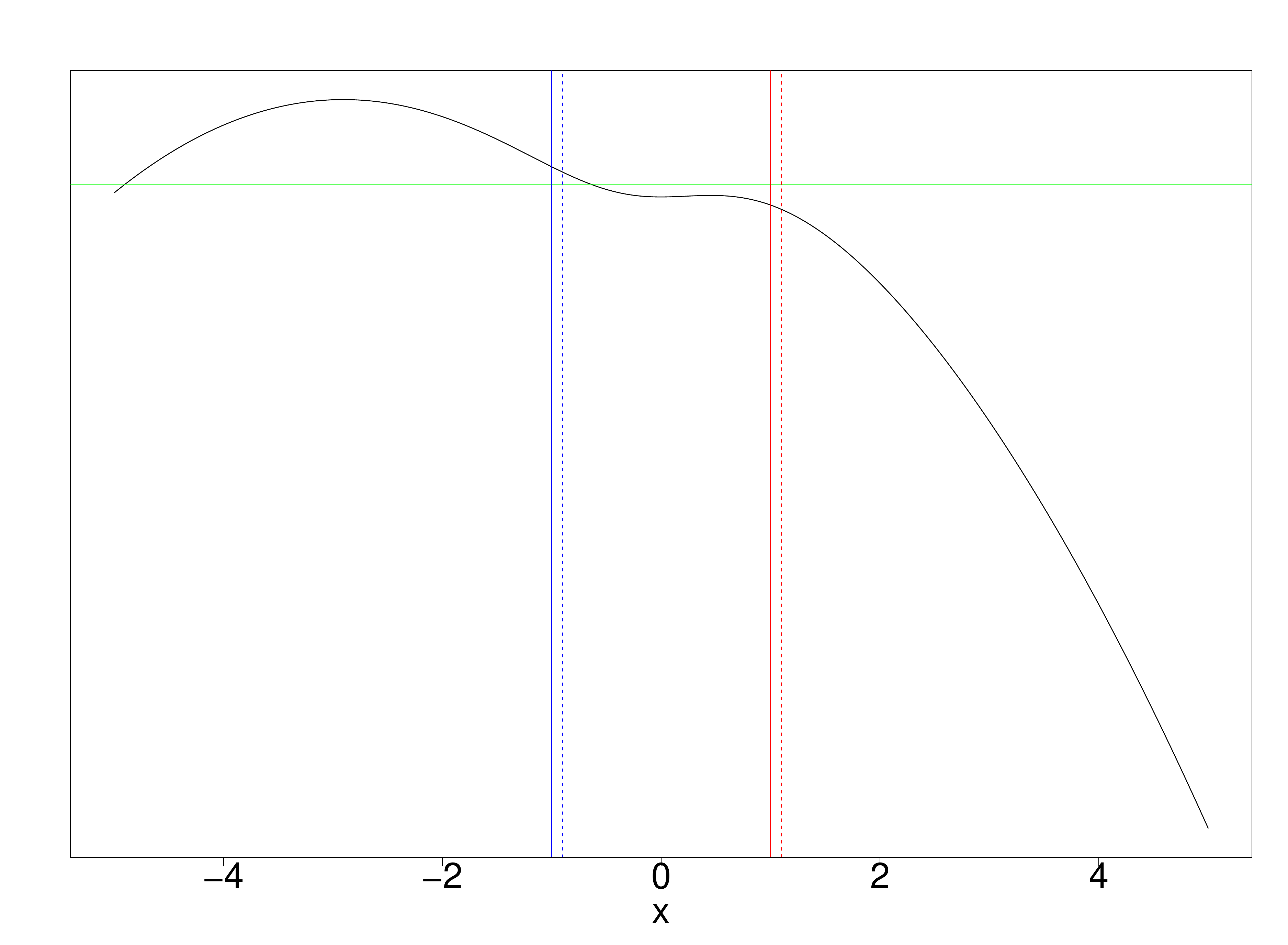}
                \caption{$\boldsymbol{\hat{\theta}} = (\hat{\mu}_1=-0.9,\hat{\mu}_2=1.1)$; \\
$\hat{\mu}_1$, $\hat{\mu}_2$ are right-shifted, $\hat{\mu}_j = {\mu}_j + 0.1$ \\}
                \label{fig:Qc for specific thetaHat cases, Example 1_case_mu1h=[-0_9]__mu2h=[1_1]}
        \end{subfigure}%

	  \begin{subfigure}[b]{0.45\textwidth}
                \centering
               \includegraphics[width=\textwidth]{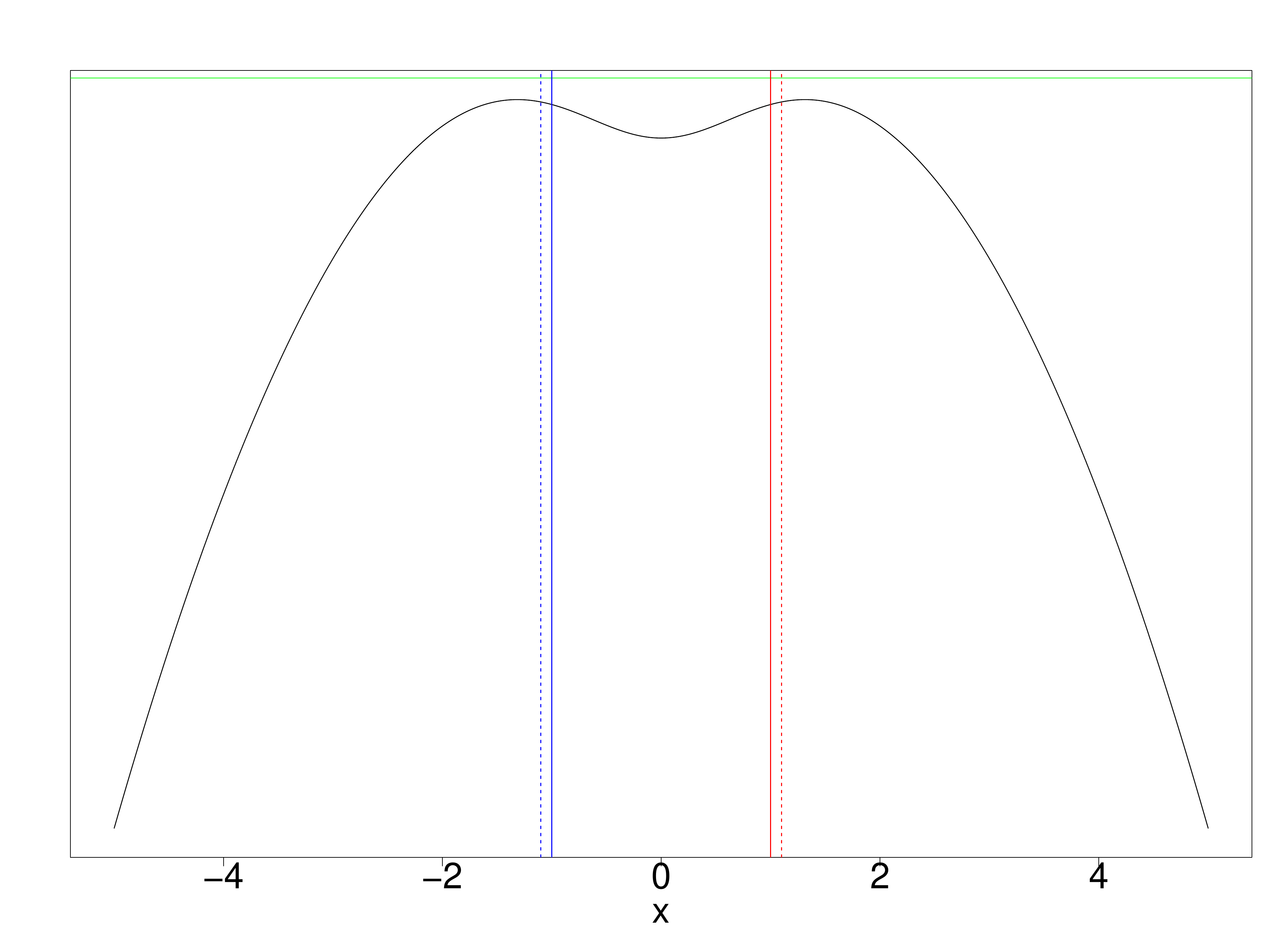}
                \caption{$\boldsymbol{\hat{\theta}} = (\hat{\mu}_1=-1.1,\hat{\mu}_2=1.1)$; \\
$\hat{\mu}_1$, $\hat{\mu}_2$ are wider, $|\hat{\mu}_j| = |{\mu}_j| + 0.1$}
                \label{fig:Qc for specific thetaHat cases, Example 1_case_mu1h=[-1_1]__mu2h=[1_1]}
        \end{subfigure}%
	  \begin{subfigure}[b]{0.45\textwidth}
                \centering
               \includegraphics[width=\textwidth]{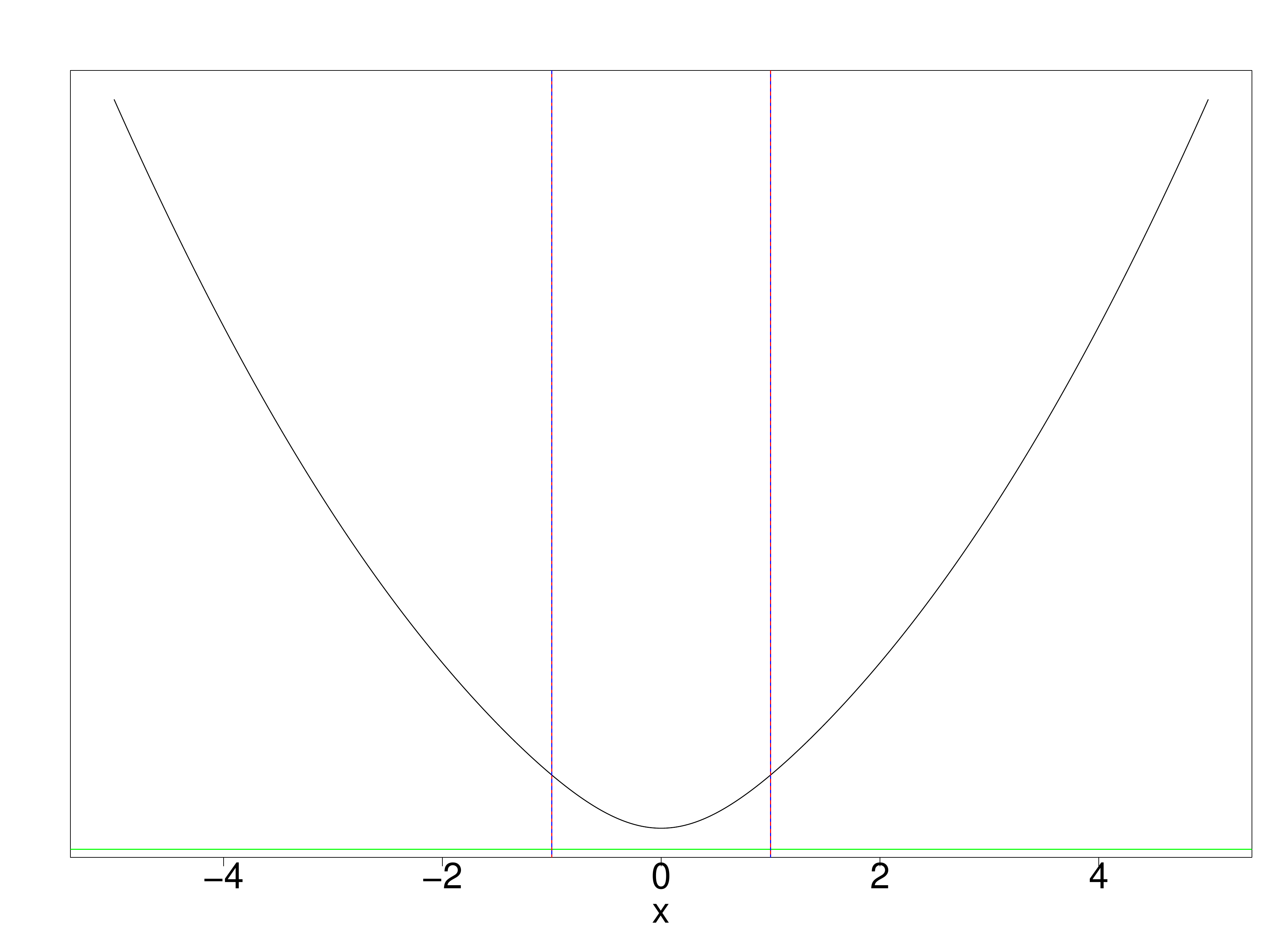}
                \caption{$\boldsymbol{\hat{\theta}} = (\hat{\mu}_1=1,\hat{\mu}_2=-1)$; \\
$\hat{\mu}_1$, $\hat{\mu}_2$ have inverse signs, $\hat{\mu}_j = -{\mu}_j$}
                \label{fig:Qc for specific thetaHat cases, Example 1_case_mu1h=[1]__mu2h=[-1]}
        \end{subfigure}%
     
        \caption{Illustration of the target $Q^c$ as a function of $x$, for specific cases of the estimated classifier parameters $\boldsymbol{\hat{\theta}} = ( \hat{\mu}_1, \hat{\mu}_2 )$.
The class mean parameters are shown in solid blue and red, with the estimated means shown in dotted blue and red. 
The green line indicates $Q^c(x) = 0$ (zero improvement), with $n_s$ being 18 in all cases.
In each specific case, the optimal selection $x_*$ yields greatest correction to $\boldsymbol{\hat{\theta}}$ in terms of moving the estimated boundary $\hat{t}$ closer to the true boundary $t$.
}
	\label{fig:Qc for specific thetaHat cases, Example 1}
\end{figure}

The target $Q^m$ is calculated analytically in Appendix A, and illustrated in Section \ref{subsubsection:The Average Performance of Shannon Entropy and Random Selection}.

\subsubsection{Exploration of Shannon Entropy and Random Selection}
\label{subsubsection:Exploration of Shannon Entropy and Random Selection}

In this theoretical example, a fully specified stochastic description of the classification problem is available, allowing the selection made by an AL method, $x_r$, to be examined and compared to the optimal selection $x_*$.
Here this comparison is made for two example selection methods, SE and RS.

SE always selects $x_r$ at the estimated boundary $\hat{t}$.
RS selects uniformly from the pool, assumed to be i.i.d. in AL, hence the RS selection probability is given by the marginal density $p(x)$.
In contrast to $Q^c$ and SE, RS is a stochastic selection method, with expected selection $x_r = 0$ in this problem.
Figure \ref{fig:Se-and-RS-and-Qc for specific thetaHat cases, Example 1} illustrates $Q^c$, SE and $p(x)$ as contrasting functions of $x$, with very different maxima.

\begin{figure}
        \centering
        \begin{subfigure}[b]{0.45\textwidth}
                \centering
               \includegraphics[width=\textwidth]{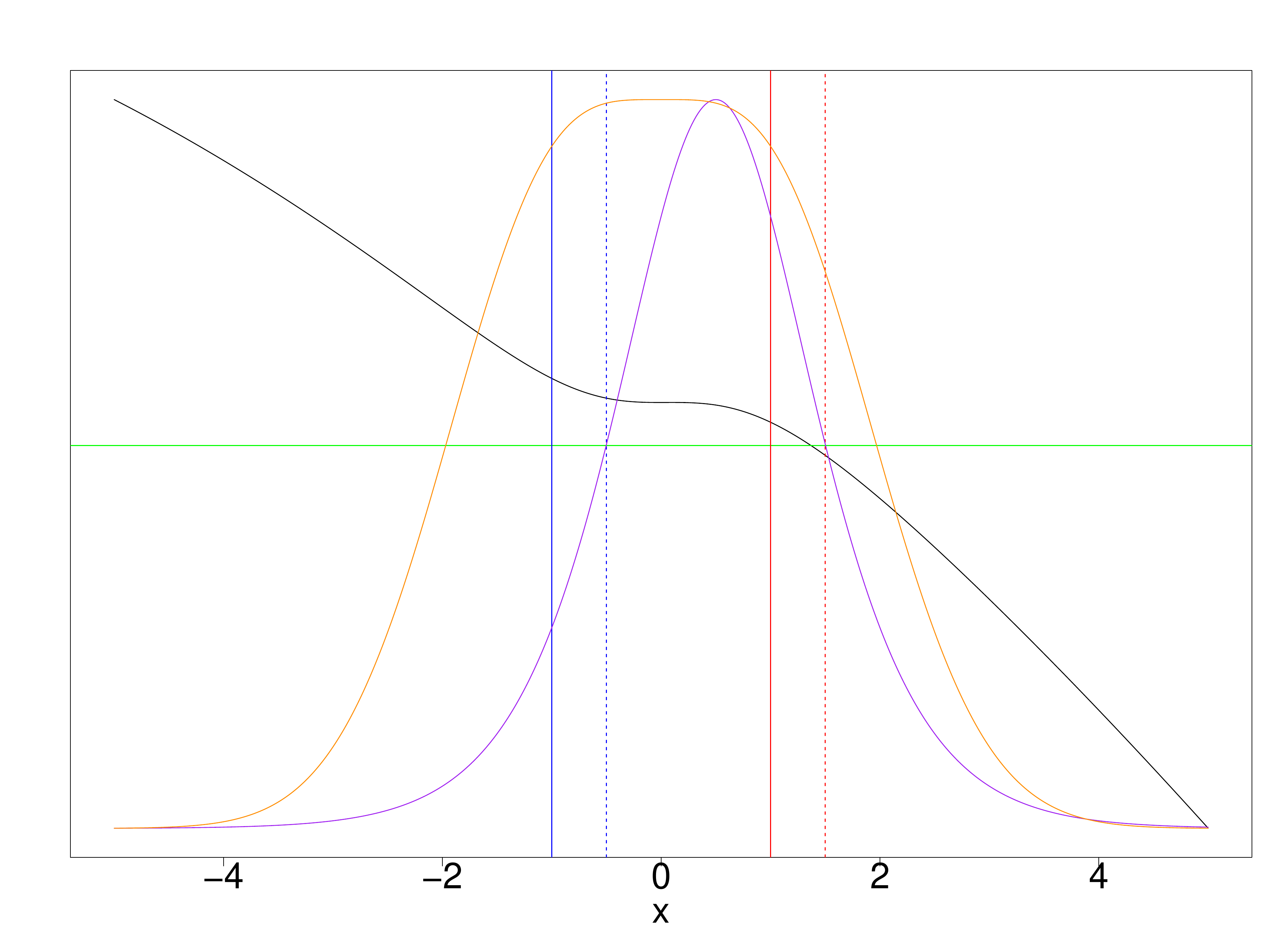}
                \caption{$\boldsymbol{\hat{\theta}} = (\hat{\mu}_1=-0.5,\hat{\mu}_2=1.5)$; \\
$\hat{\mu}_1$, $\hat{\mu}_2$ are right-shifted, $\hat{\mu}_j = {\mu}_j + 0.5$ \\}
                \label{fig:Se-and-Qc for specific thetaHat cases, Example 1_case_mu1h=[-0_5]__mu2h=[1_5]}
        \end{subfigure}
        \begin{subfigure}[b]{0.45\textwidth}
                \centering
               \includegraphics[width=\textwidth]{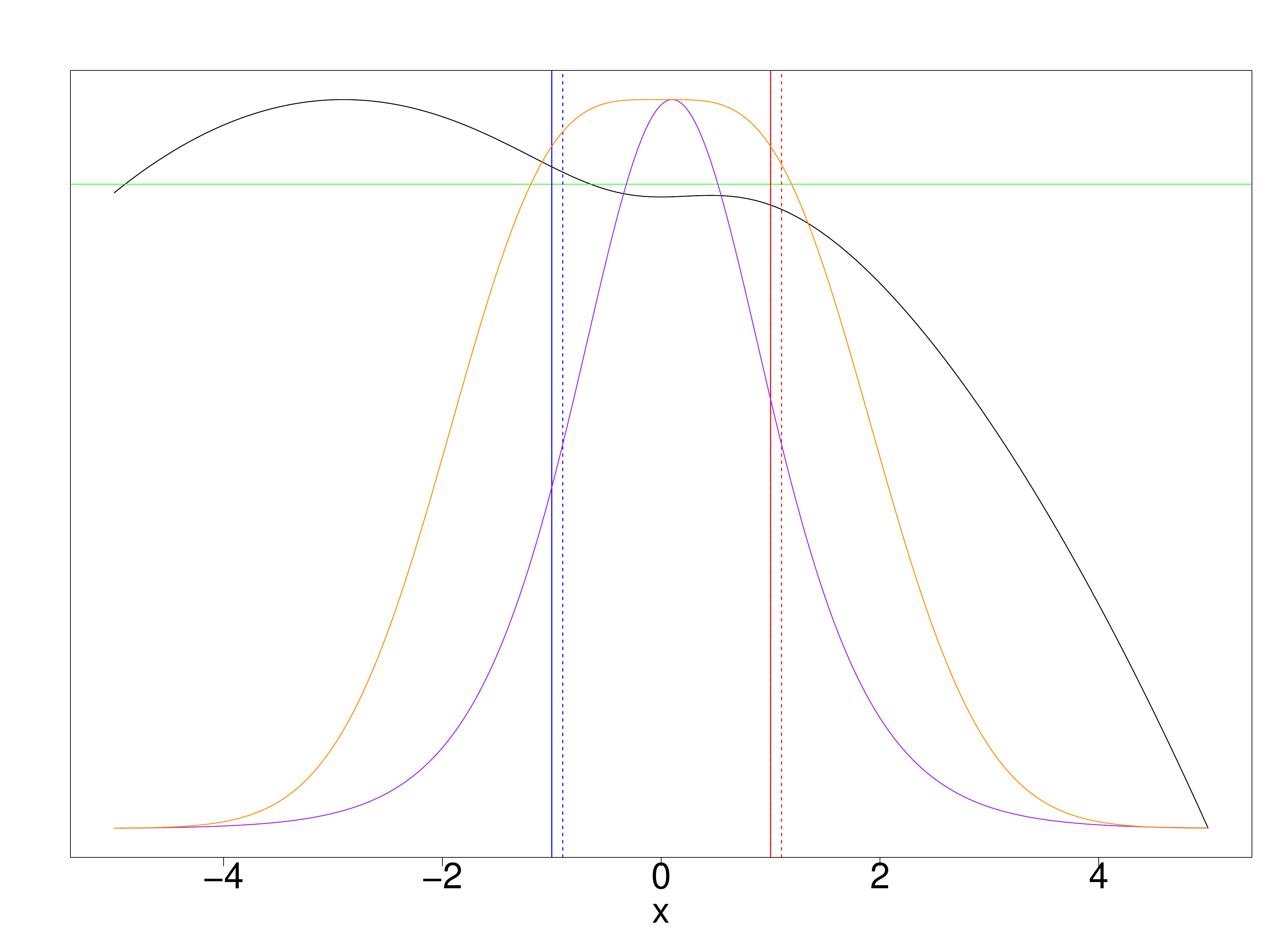}
                \caption{$\boldsymbol{\hat{\theta}} = (\hat{\mu}_1=-0.9,\hat{\mu}_2=1.1)$; \\
$\hat{\mu}_1$, $\hat{\mu}_2$ are right-shifted, $\hat{\mu}_j = {\mu}_j + 0.1$ \\}
                \label{fig:Se-and-Qc for specific thetaHat cases, Example 1_case_mu1h=[-0_9]__mu2h=[1_1]}
        \end{subfigure}%

	  \begin{subfigure}[b]{0.45\textwidth}
                \centering
               \includegraphics[width=\textwidth]{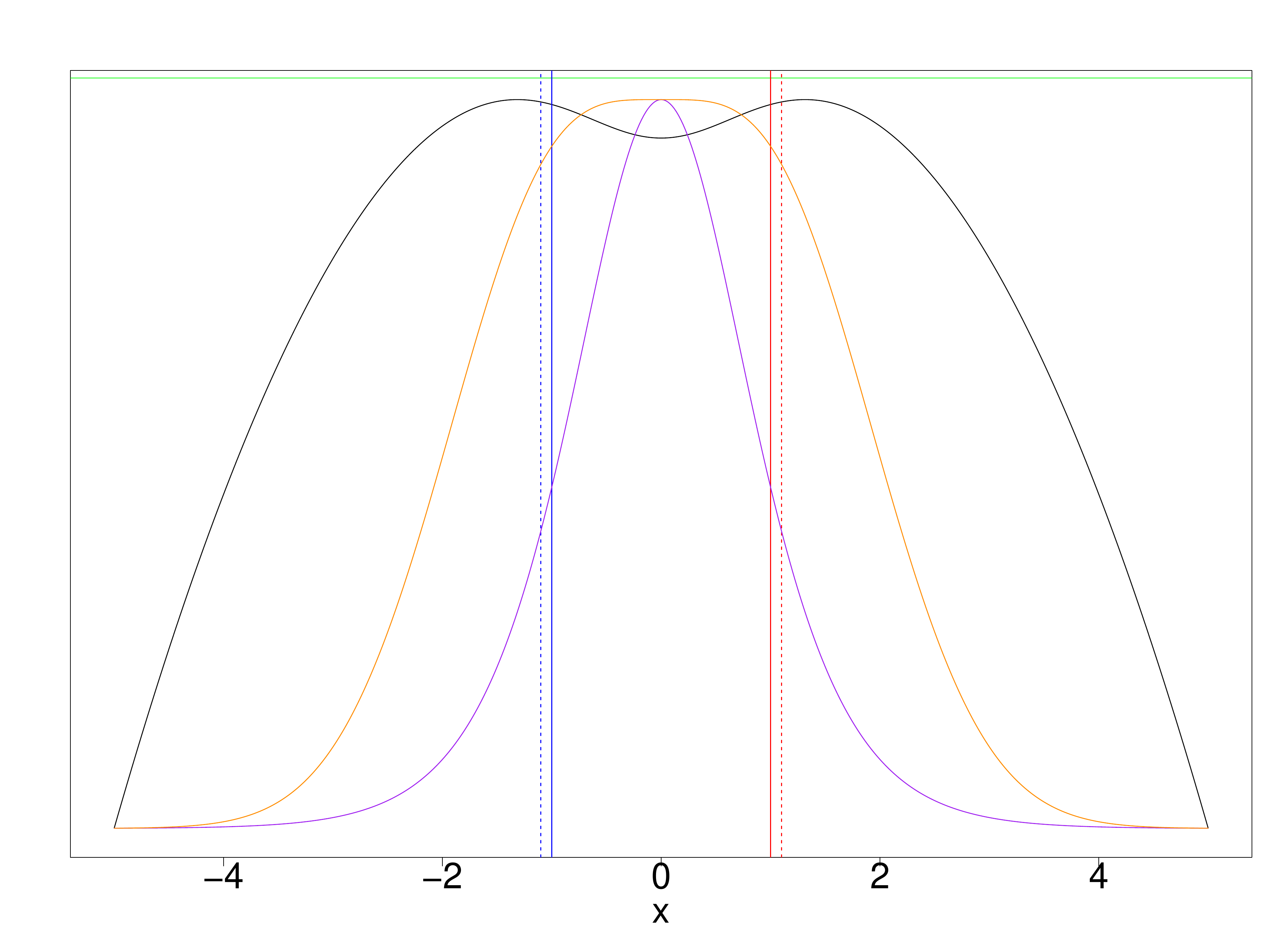}
                \caption{$\boldsymbol{\hat{\theta}} = (\hat{\mu}_1=-1.1,\hat{\mu}_2=1.1)$; \\
$\hat{\mu}_1$, $\hat{\mu}_2$ are wider, $|\hat{\mu}_j| = |{\mu}_j| + 0.1$}
                \label{fig:Se-and-Qc for specific thetaHat cases, Example 1_case_mu1h=[-1_1]__mu2h=[1_1]}
        \end{subfigure}%
	  \begin{subfigure}[b]{0.45\textwidth}
                \centering
               \includegraphics[width=\textwidth]{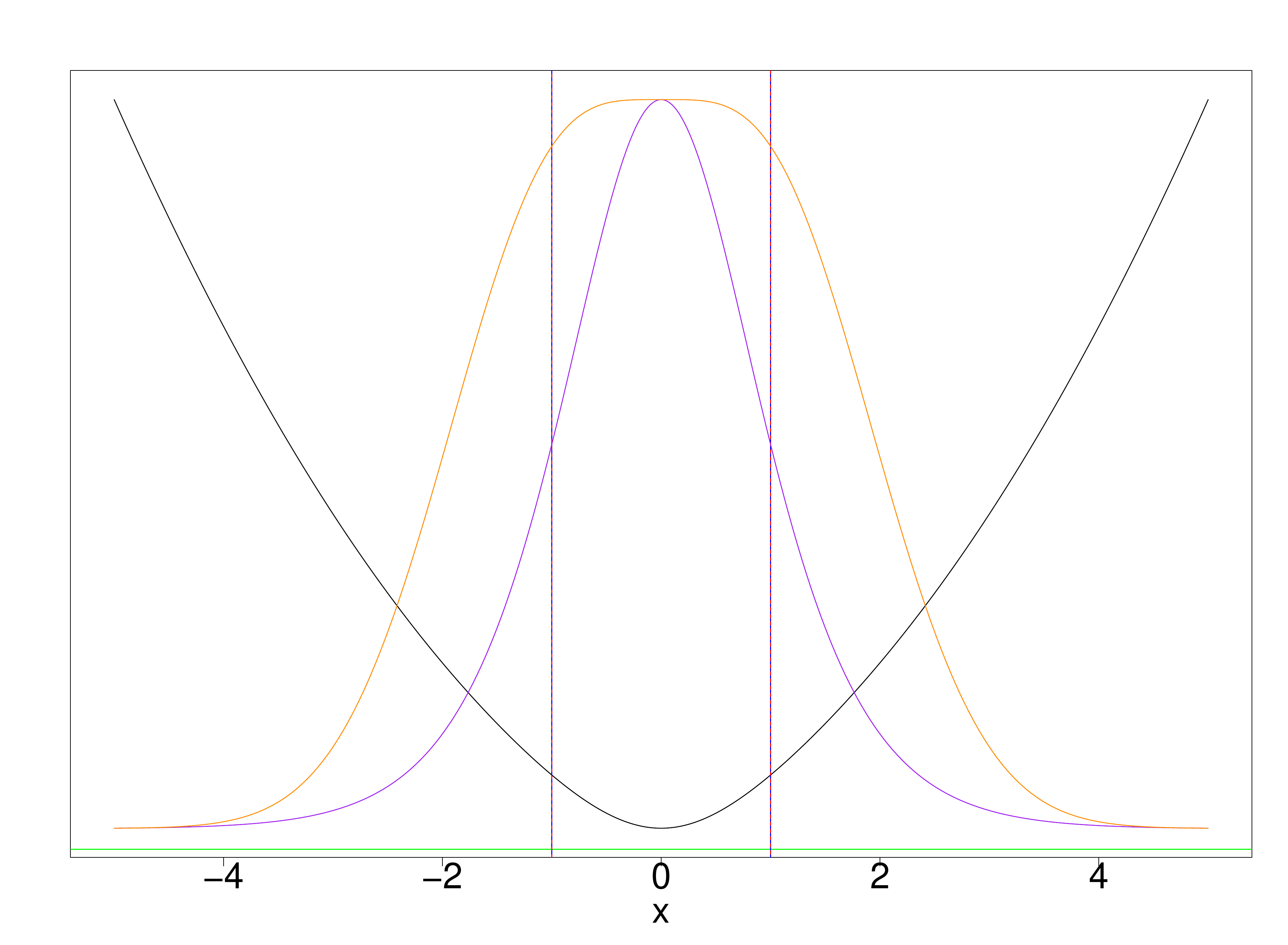}
                \caption{$\boldsymbol{\hat{\theta}} = (\hat{\mu}_1=1,\hat{\mu}_2=-1)$; \\
$\hat{\mu}_1$, $\hat{\mu}_2$ have inverse signs, $\hat{\mu}_j = -{\mu}_j$}
                \label{fig:Se-and-Qc for specific thetaHat cases, Example 1_case_mu1h=[1]__mu2h=[-1]}
        \end{subfigure}%

        \caption{Comparison of $Q^c$ against SE and RS as functions of $x$, for specific cases of the estimated classifier parameters 
$\boldsymbol{\hat{\theta}} = ( \hat{\mu}_1, \hat{\mu}_2 )$.
$Q^c$ is shown in black, SE in purple and RS in orange (for RS, the density $p(x)$ is shown).
The class mean parameters are shown in solid blue and red, with the estimated means shown in dotted blue and red. 
The green line indicates $Q^c(x) = 0$ (zero improvement), with $n_s$ being 18 in all cases.
The three functions are scaled to permit this comparison.
}
	\label{fig:Se-and-RS-and-Qc for specific thetaHat cases, Example 1}
\end{figure}

$Q^c$ is asymmetric in the first two cases, and symmetric for the final two. 
By contrast, SE and RS are always symmetric (for all possible values of $\boldsymbol{\hat{\theta}}$).

In the first two cases (Figures \ref{fig:Se-and-Qc for specific thetaHat cases, Example 1_case_mu1h=[-0_5]__mu2h=[1_5]} and \ref{fig:Qc for specific thetaHat cases, Example 1_case_mu1h=[-0_9]__mu2h=[1_1]}), SE selects a central $x_r$, thereby missing the optimal selection $x_*$. 
In the second case (Figure \ref{fig:Qc for specific thetaHat cases, Example 1_case_mu1h=[-0_9]__mu2h=[1_1]}), SE selects $x_r$ with $Q^c(x_r) < 0$, failing to improve the classifier, whereas the optimal selection $x_*$ does improve the classifier since $Q^c(x_*) > 0$.
The third case is unusual, since $\hat{t} = t$ and this classifier's loss $L_e$ cannot be improved, hence $Q^c(x) < 0$ for all $x$.
In the fourth case (Figure \ref{fig:Se-and-Qc for specific thetaHat cases, Example 1_case_mu1h=[1]__mu2h=[-1]}) SE makes the worst possible choice of $x$.
In all four cases, SE never chooses the optimal point; SE may improve the classifier, but never yields the greatest improvement.
These specific cases of $\boldsymbol{\hat{\theta}}$ show that SE often makes a suboptimal choice for $x_r$, for this theoretical example.

Turning to consider RS, for these four cases of $\boldsymbol{\hat{\theta}}$, the expected RS selection is a suboptimal choice of $x_*$ for this problem.
It is notable that the expected RS selection is usually close to the SE selection.
The stochastic nature of RS implies that it often selects far more non-central $x$ values than SE.


\subsubsection{The Average Performance of Shannon Entropy and Random Selection}
\label{subsubsection:The Average Performance of Shannon Entropy and Random Selection}

Having explored specific cases of $\boldsymbol{\hat{\theta}}$, the performance of AL methods on average over the labelled data $D_S$ is now considered, by the comparison of SE and RS to the target $Q^m$.
$Q^m$ illuminates optimal AL selection for the classification problem itself, independently of any specific dataset $D_S$.

\begin{figure}
        \centering
        \begin{subfigure}[b]{0.45\textwidth}
                \centering
               \includegraphics[width=\textwidth]{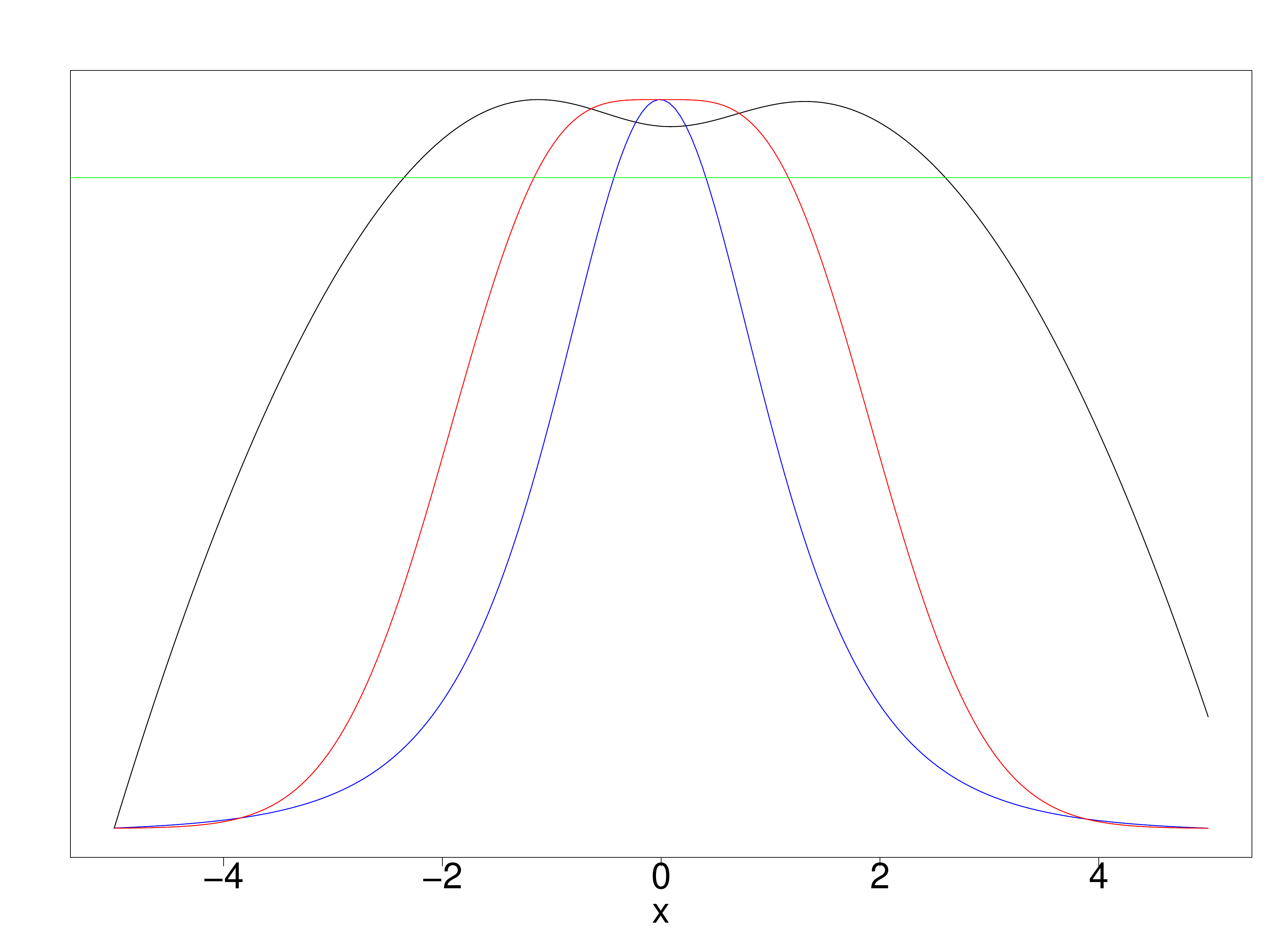}
                \caption{Problem A: $ \{ {\boldsymbol\pi} = (\frac{1}{2}, \frac{1}{2}), \\
(X|Y=c_1) \sim \textrm{N}(-1,1), \\
(X|Y=c_2) \sim \textrm{N}(1,1) \}$}
                \label{fig:SE-and-RS-and-Qm for Example 1}
        \end{subfigure}
        \begin{subfigure}[b]{0.45\textwidth}
                \centering
               \includegraphics[width=\textwidth]{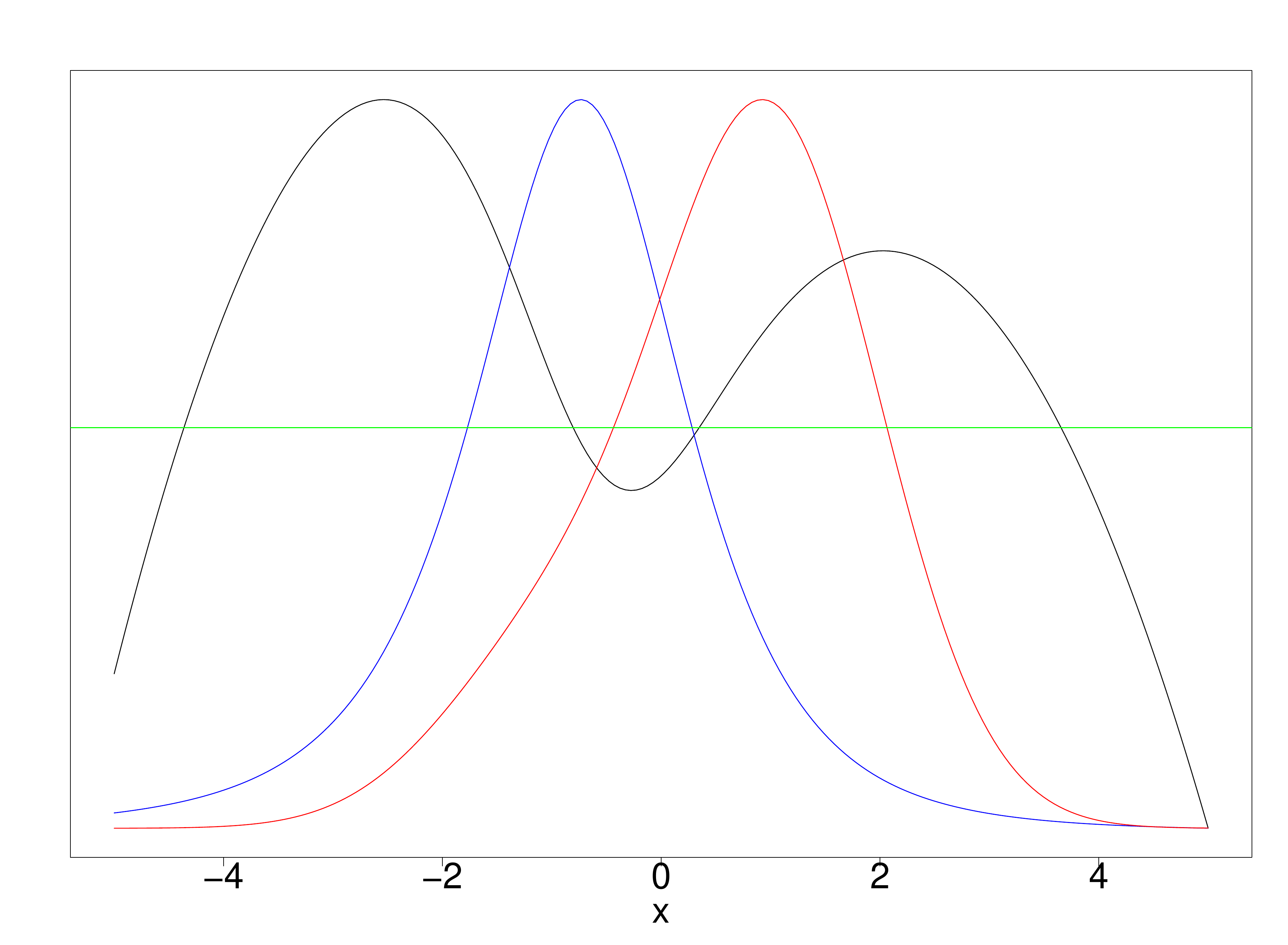}
                \caption{Problem B: $ \{ {\boldsymbol\pi} = (\frac{1}{5}, \frac{4}{5}), \\
(X|Y=c_1) \sim \textrm{N}(-1,1), \\
(X|Y=c_2) \sim \textrm{N}(1,1) \}$}
                \label{fig:SE-and-RS-and-Qm for Example 2}
        \end{subfigure}%


        \caption{Comparison of SE and RS against $Q^m$ as functions of $x$, with two closely related theoretical problems, which differ only in the class prior ${\boldsymbol{\pi}}$.
For both problems there are large ranges of the covariate $x$ which improve the classifier, in expectation over $D_S$.
$Q^m$ is shown in black, SE in blue, and RS in red (for RS, the density $p(x)$ is shown).
The green line indicates $Q^m(x)= 0$ (zero improvement), with $n_s$ being 18 in both cases.
The three functions are scaled to permit this comparison.
}
	\label{fig:Se-and-Rs-and-Qm for two Examples}
\end{figure}

This comparison is made with two theoretical problems, the first being defined at the start of Section \ref{subsection:Theoretical Example to Illustrate Example Quality}, and the second problem being a minor variant where the prior $\boldsymbol{\pi}$ has been modified to $(\frac{1}{5}, \frac{4}{5})$.
Both problems are defined in Figure \ref{fig:Se-and-Rs-and-Qm for two Examples}.
The values for $Q^m$ and SE are calculated by averaging over multiple i.i.d. draws of the labelled data $D_S$.

Figure \ref{fig:Se-and-Rs-and-Qm for two Examples} illustrates the comparison of SE and RS against $Q^m$ with these two theoretical problems.
This figure suggests that for both problems, there are large ranges of the covariate $x$ which improve the classifier, in expectation over $D_S$.
The locations of the improvement $x$ values are quite different for the two problems.

\subsection{The Dependence of Active Learning on the Labelled Data}
\label{subsection:The Relationship between Marginal EQ and Conditional EQ}

\def \Qc_RankPlots1_Textwidth_Multiplier {0.48}


\begin{figure}
        \centering
        \begin{subfigure}[b]{\Qc_RankPlots1_Textwidth_Multiplier \textwidth}
                \centering
               \includegraphics[width=\textwidth]{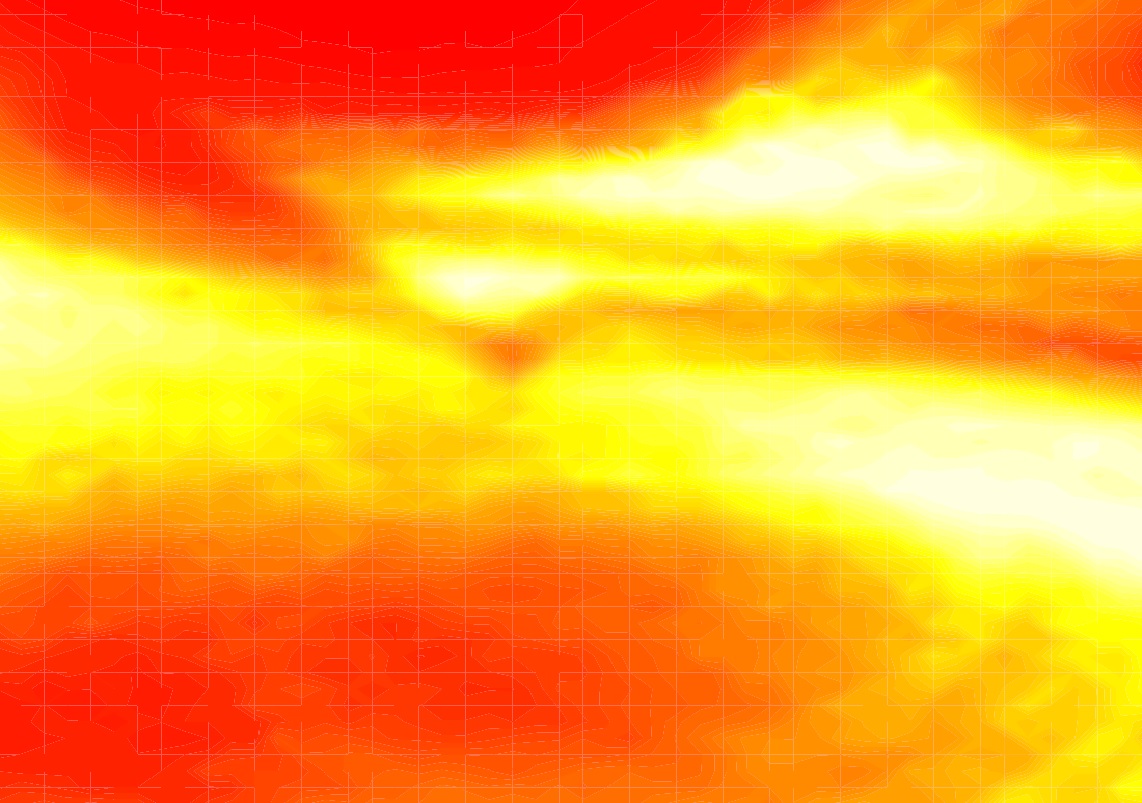}
                \caption{$Q^c$ ranks given $D_1$}
                \label{fig:Qc rankings of Ripley to show similarity of rankings 1___ranksd1}
        \end{subfigure}
        \begin{subfigure}[b]{\Qc_RankPlots1_Textwidth_Multiplier \textwidth}
                \centering
               \includegraphics[width=\textwidth]{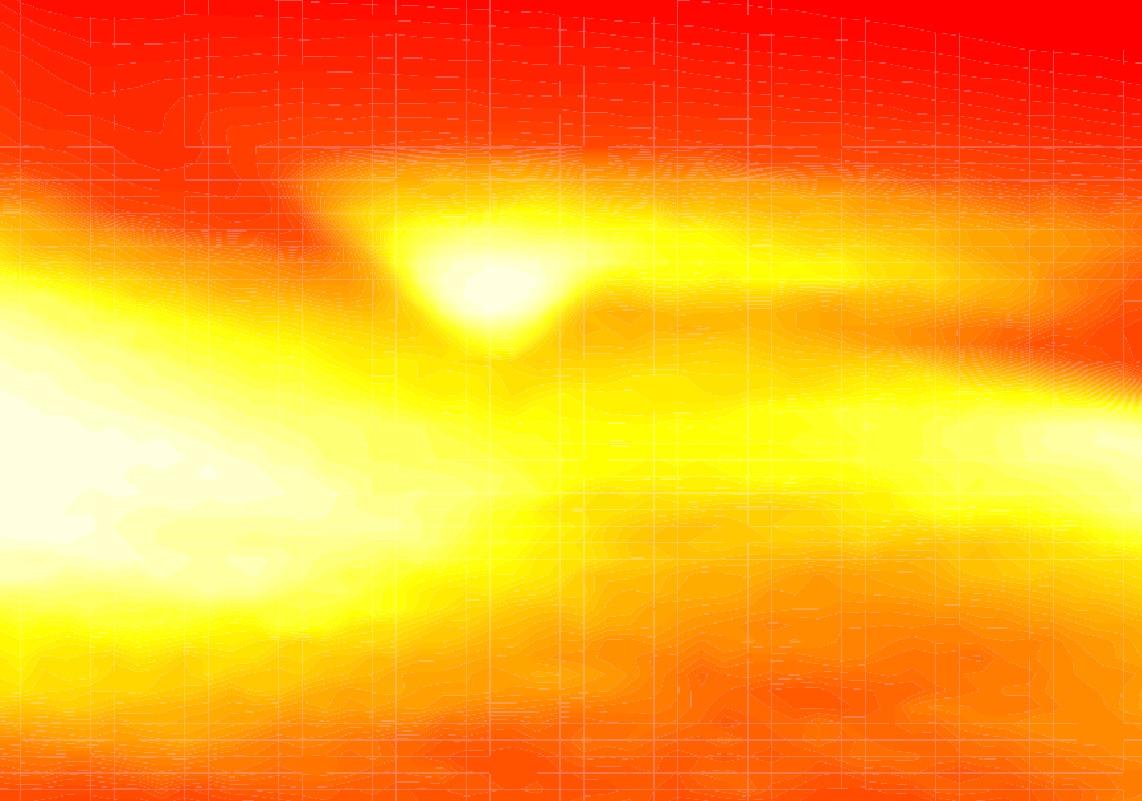}
                \caption{$Q^c$ ranks given $D_2$}
		   \label{fig:Qc rankings of Ripley to show similarity of rankings 1___ranksd2}
        \end{subfigure}

        \begin{subfigure}[b]{\Qc_RankPlots1_Textwidth_Multiplier \textwidth}
                \centering
               \includegraphics[width=\textwidth]{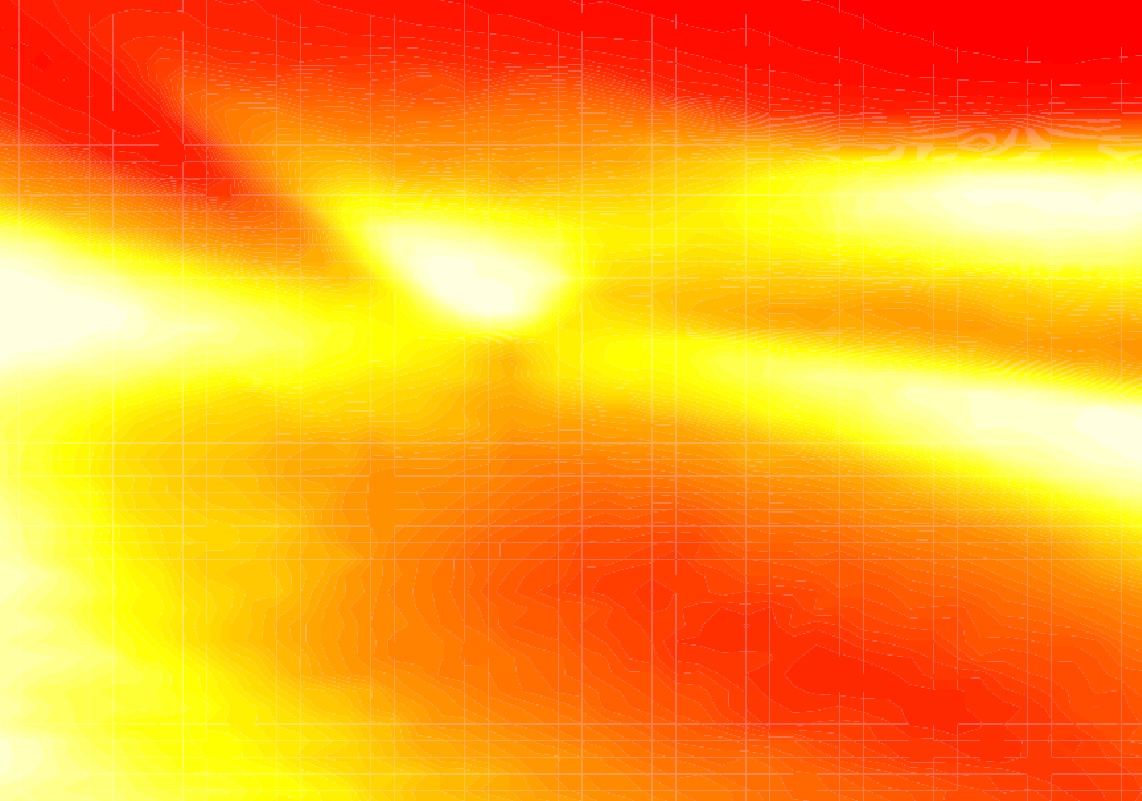}
                \caption{$Q^c$ ranks given $D_3$}
		   \label{fig:Qc rankings of Ripley to show similarity of rankings 1___ranksd3}
        \end{subfigure}
        \begin{subfigure}[b]{\Qc_RankPlots1_Textwidth_Multiplier \textwidth}
                \centering
               \includegraphics[width=\textwidth]{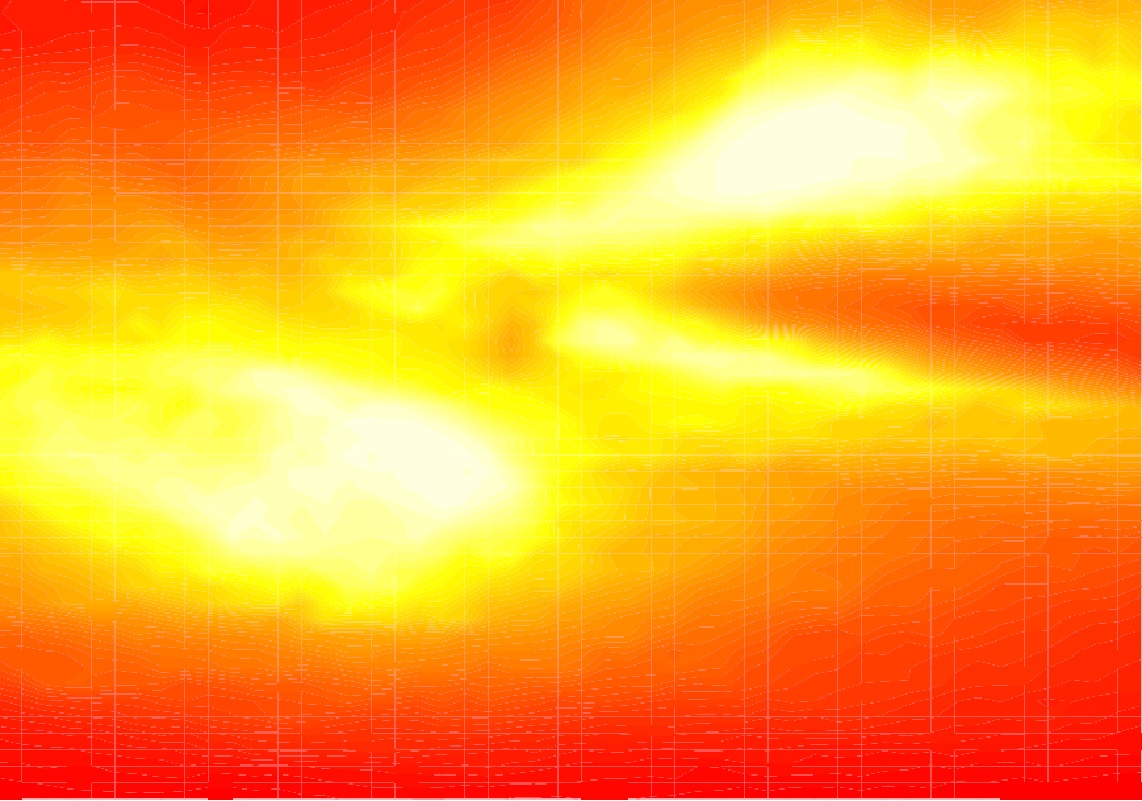}
                \caption{$Q^c$ ranks given $D_4$}
		   \label{fig:Qc rankings of Ripley to show similarity of rankings 1___ranksd4}
        \end{subfigure}

        \begin{subfigure}[b]{\Qc_RankPlots1_Textwidth_Multiplier \textwidth}
                \centering
                \includegraphics[width=\textwidth]{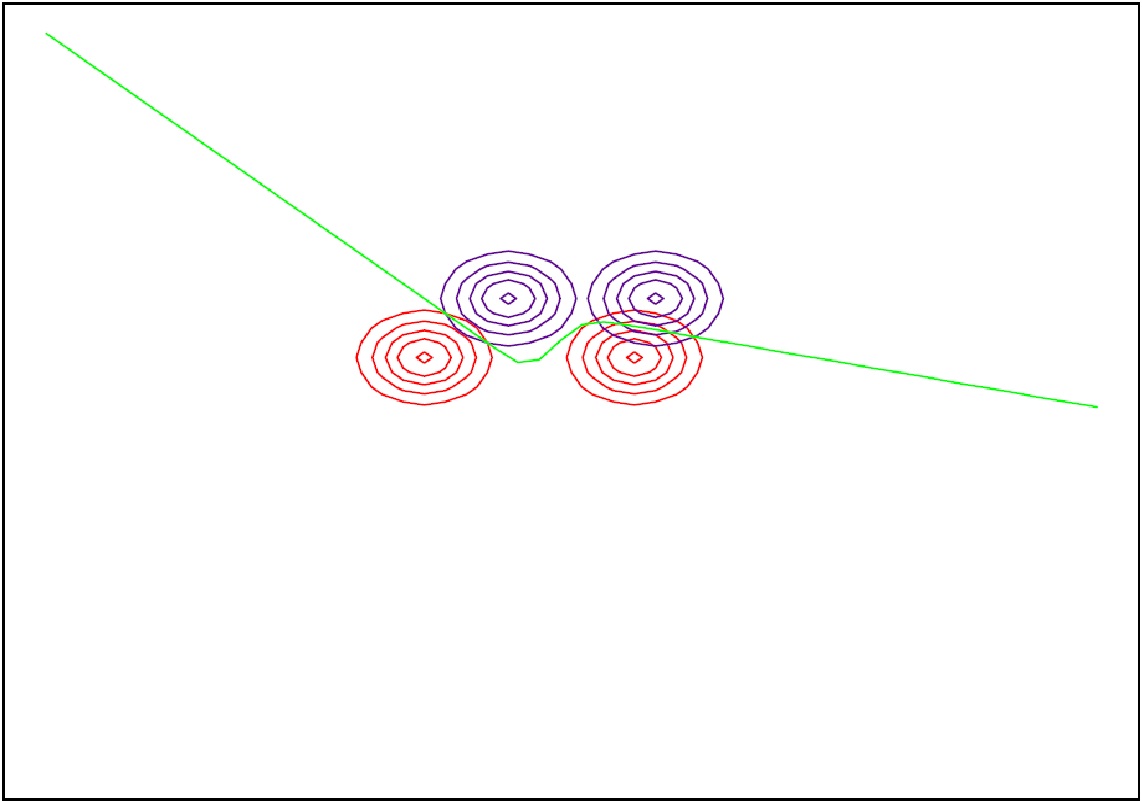}
                \caption{Stochastic Truth: \textcolor{red}{$X|Y=c_1$}, \textcolor{ao}{$X|Y=c_2$}}
		   \label{fig:Qc rankings of Ripley to show similarity of rankings 1___Truth}
        \end{subfigure}
        \begin{subfigure}[b]{\Qc_RankPlots1_Textwidth_Multiplier \textwidth}
                \centering
                \includegraphics[width=\textwidth]{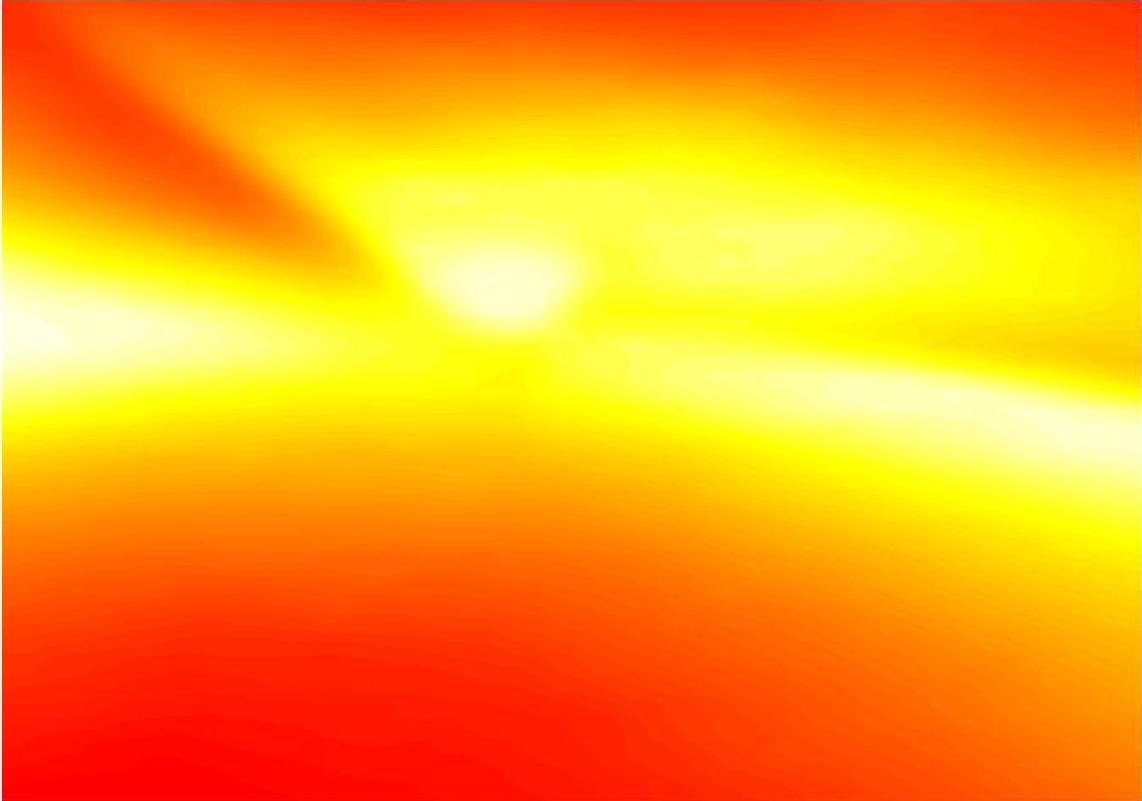}
                \caption{Averaged $Q^c$ ranks}
		   \label{fig:Qc rankings of Ripley to show similarity of rankings 1___averagedRanks}
        \end{subfigure}
     
        \caption{
The visual similarity of the four $Q^c$ rankings (Figures \ref{fig:Qc rankings of Ripley to show similarity of rankings 1___ranksd1} to \ref{fig:Qc rankings of Ripley to show similarity of rankings 1___ranksd4}) show the low dependence of AL on the labelled data, in other words the low sensitivity of $Q^c$ to $D_S$.
Ranking of $Q^c$ values for different draws of the fixed-size dataset $D_i$ are shown in Figures \ref{fig:Qc rankings of Ripley to show similarity of rankings 1___ranksd1} to \ref{fig:Qc rankings of Ripley to show similarity of rankings 1___ranksd4}.
Higher ranks are shown in brighter yellow, lower ranks are darker red. 
The classification problem is the Ripley four-Gaussian problem, where the stochastic truth in shown in Figure \ref{fig:Qc rankings of Ripley to show similarity of rankings 1___Truth}, see Appendix B.
The size of $D_i$ is always 20.
}
	\label{fig:Qc rankings of Ripley to show similarity of rankings 1}
\end{figure}


Most if not all AL methods implicitly condition on the labelled data $D_S$, via dependence on the classifier $\boldsymbol{\hat{\theta}}$, for example uncertainty sampling and QBC (described in Section \ref{subsection:Literature Review}).
The improvement function $Q^c$ and its maximum $x_*$ also condition on $D_S$, see Equation \ref{eq:eqc}.
This conditioning on $D_S$ is very natural in AL, yet is rarely made explicit, or examined.
This dependency on $D_S$ first motivates the definition of $Q^m$ in Equation \ref{eq:eqm}, which addresses the dependence taking the expectation over $D_S$.
However the dependence on $D_S$ also raises another statistical question: how does optimal AL selection depend on the labelled data $D_S$?

To illustrate, one possibility is that $Q^c$ and $x_*$ vary wildly with $D_S$; in this case (denoted \emph{fragile}), different draws of $D_S$ would lead to very different $x_*$ values.
Alternatively $x_*$ might have low sensitivity to $D_S$, and in this second case (denoted \emph{robust}), different draws of $D_S$ would produce similar $x_*$ values.
Figure \ref{fig:Qc rankings of Ripley to show similarity of rankings 1} shows $Q^c$ and $x_*$ for different draws of $D_S$, where the similarity of Figures \ref{fig:Qc rankings of Ripley to show similarity of rankings 1___ranksd1} to \ref{fig:Qc rankings of Ripley to show similarity of rankings 1___ranksd4} and \ref{fig:Qc rankings of Ripley to show similarity of rankings 1___averagedRanks} show the low sensitivity of $Q^c$ to $D_S$.


Each draw of $D_S$ provides a classifier $\boldsymbol{\hat{\theta}}$, which in turn provides a specific $Q^c$ ranking of all the examples in the pool $X_P$, shown in Figure \ref{fig:Qc rankings of Ripley to show similarity of rankings 1}.
In the second robust case these pool rankings will be similar to each other, and this is observed visually 
in Figure \ref{fig:Qc rankings of Ripley to show similarity of rankings 1}.
To examine this statistically, experiments examine the sensitivity of $Q^c$ to $D_S$.
The similarity of the pool rankings from different draws of $D_S$ express the sensitivity of $Q^c$ to $D_S$, where greater similarity shows lower sensitivity.

Here a different theoretical classification problem is considered, the Ripley four-\-Gaussian problem (shown in Figure \ref{fig:Qc rankings of Ripley to show similarity of rankings 1___Truth}).
The classifier is quadratic discriminant analysis \citep[Chapter~4]{Tibshirani2009}, with the loss function being error rate $L_e$.

A single constant pool of unlabelled examples $X_P$ is chosen as a fixed grid of points in covariate space, with a 2-d grid providing visualisation in Figure \ref{fig:Qc rankings of Ripley to show similarity of rankings 1}.
Multiple datasets $(D_i)_1^n$ are drawn i.i.d. from $({\bf X}, Y)$.
Each dataset $D_i$ produces a set of $Q^c$ values for the grid, denoted ${\bf q}_i$, implying a ranking of the pool, denoted ${\bf r}_i$, shown in Figure \ref{fig:Qc rankings of Ripley to show similarity of rankings 1}.

The similarity of the pool rankings $({\bf r}_i)_1^n$ is examined by standard statistical tests of correlations (with Holm and Bonferroni corrections), Moran's I and Geary's C \citep{Moran1950,Geary1954}.
These statistical tests show that the ranks are very closely related for different draws of $(D_i)_1^n$.

This similarity of the ranks shows that, in this toy example, $Q^c$ has low dependence on the specific dataset $D_S$, which is the robust case.
This in turn suggests that the optimal AL choice $x_*$ for one dataset $D_i$ is near-optimal for a different dataset $D_j$.

\section{Algorithms to Estimate Example Quality}
\label{section:Algorithms to Estimate Example Quality}

Section \ref{section:Example Quality} defines the targets $Q^c$ and $Q^m$ in a theoretical context with a full stochastic description of the classification problem.
AL applications present a very different context, where all theoretical quantities must be estimated from the single labelled dataset $D_S$.
In AL applications, there is only one label budget, and as a result AL methods cannot be compared to each other, or to benchmarks such as RS \citep{Provost2010}.
The estimation of multiple theoretical quantities from the single dataset $D_S$ introduces statistical issues described below.


The definition of $Q^c$ in Equation \ref{eq:eqc} includes three primary components: ${\bf p}$, $\boldsymbol{\hat{\theta}}$ and $L$.
$Q^c$ estimation requires estimating these three components from one labelled dataset $D_S$.
Estimating three quantities from a single dataset raises interesting statistical choices.
One major choice must be made between using the same data for all three estimations, termed \emph{na\"ive reuse}, or partitioning the data into subsets for each estimation.
This choice between na\"ive reuse and partitioning has implications for the bias and variance of $\hat{Q}^c$ estimates, discussed below.

Notationally three datasets are denoted $D_C$, $D_T$ and $D_E$ to estimate three components of $Q^c$:
\begin{itemize}
\item The class probability vector, ${\bf p} = p(Y|{\bf x})$, estimated by $\hat{\bf p}$ using data $D_C$,
\item The classifier parameters, $\boldsymbol{\hat{\theta}} = \theta(D_S)$, estimated by $\theta(D_T)$ using data $D_T$,
\item The loss, $L$, estimated by $\hat{L}$ using data $D_E$.
\end{itemize}
Each of these three datasets must be subsets of $D_S$.

In the case of na\"ive reuse, all three datasets equal $D_S$.
For data partitioning, the three datasets are disjoint partitions of $D_S$.
Both of these cases are shown in Figure \ref{fig:Diagram to show two Qc estimating algorithms}.
More sophisticated partitioning schemes are under investigation and the subject of future work.

\subsection{Two EQ Estimation Algorithms}
\label{subsection:Algorithms of Different Estimation Bias}

\begin{figure}
\centering
\includegraphics[scale=0.4]{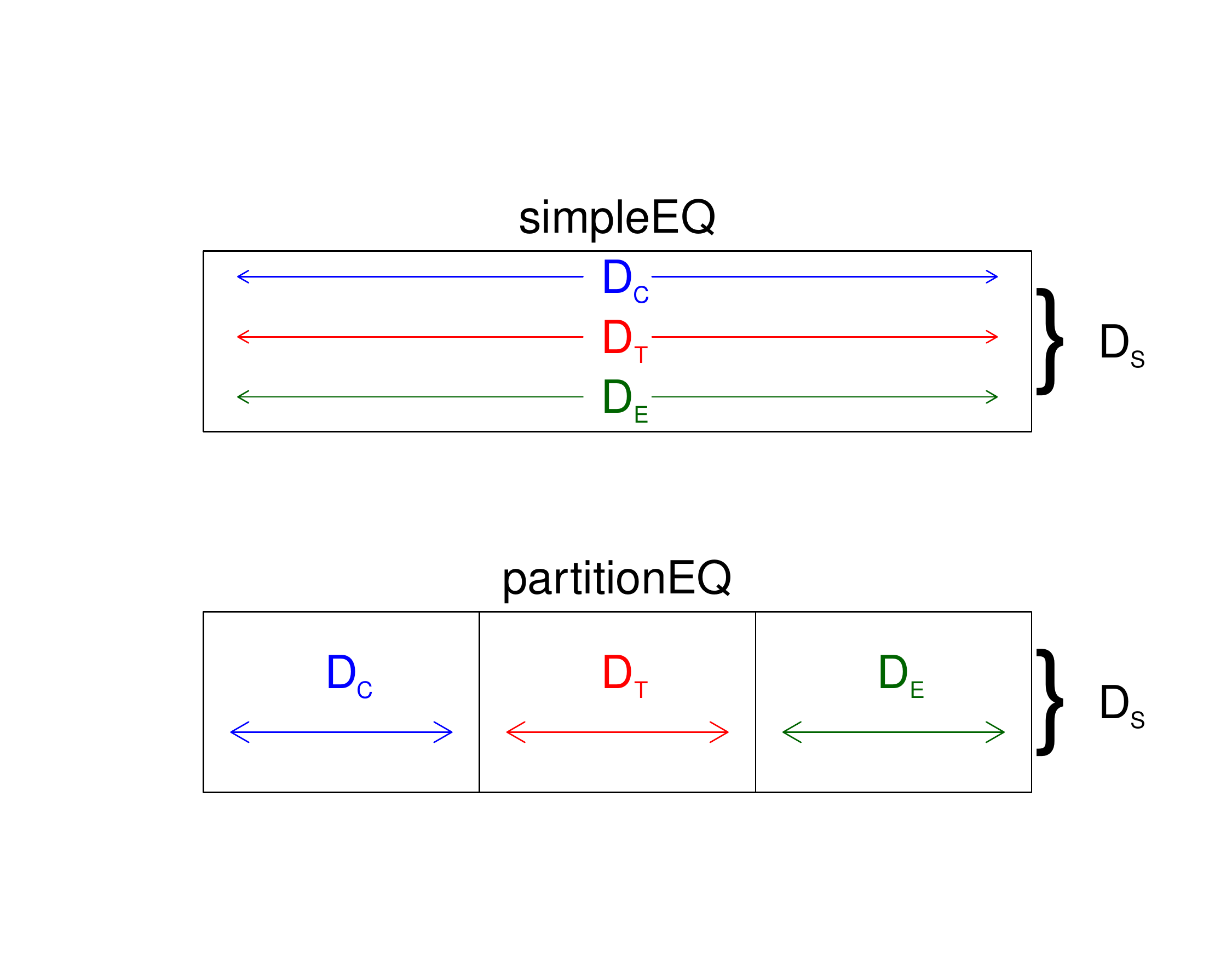}
\caption{Different usage of the labelled data $D_S$ by two $Q^c$-estimation algorithms.
SimpleEQ uses all of $D_S$ to estimate all three components; this is na\"ive reuse.
PartitionEQ divides $D_S$ into disjoint subsets, then uses one subset per component; this is data partitioning.
}
\label{fig:Diagram to show two Qc estimating algorithms}
\end{figure}

Two algorithms are presented that estimate $Q^c$ directly.
The first algorithm takes a simple approach where all of $D_S$ is used to estimate all three components.
The intention is to reduce the variance of the component estimates, by using the maximum amount of data for each estimate.
This first algorithm is termed simpleEQ; its na\"ive reuse of the data is illustrated in Figure \ref{fig:Diagram to show two Qc estimating algorithms}.
Here $D_S = D_C = D_T = D_E$.

For practical estimation of $Q^c$, Term F in Equation \ref{eq:eqc} can be ignored since it is independent of ${\bf x}$.
Thus the main task of practical $Q^c$ estimation is estimation of Term J in Equation \ref{eq:eqc}, Term J being the expected classifier loss after retraining on the new example ${\bf x}$ with its unknown label $Y|{\bf x}$.

The simpleEQ algorithm immediately encounters a problem in estimating Term J: the same data $D_S$ is used both to train the classifier and also to estimate the loss.
This in-sample loss estimation is known to produce optimistic, biased estimates of the loss \citep[Chapter~7]{Tibshirani2009}.

This problem of biased component estimates under na\"ive reuse motivates the development of a second algorithm, termed partitionEQ.
The intention is to produce unbiased component estimates, and thereby reduce the bias of the $\hat{Q}^c$ estimate.
This algorithm randomly partitions the data $D_S$ into three disjoint subsets $D_C$, $D_T$ and $D_E$; this partitioning is shown in Figure \ref{fig:Diagram to show two Qc estimating algorithms}.

Each subset is used for a single estimation task: to estimate ${\bf p}$, the classifier parameters $\boldsymbol{\hat{\theta}}$, and the loss $L$ respectively.
The random partitioning of $D_S$ into subsets is arbitrary.
For this reason, the random partitioning is performed several times, and the resulting $\hat{Q}^c$ estimates are averaged.
In the experimental study of Section \ref{section:Experiments and Results}, the partitioning is repeated ten-fold.

Random sub-sampling of the pool is used for computational efficiency.

\section{Experiments and Results}
\label{section:Experiments and Results}

A large-scale experimental study explores the performance of the new $Q^c$-based AL methods.
The intention is to compare those methods with each other, and to standard AL methods from the literature (described in Section \ref{subsection:Literature Review}).
The focus is on the relative classifier improvements of each AL method, rather than absolute classifier performance.

The base classifier is varied, since AL performance is known to depend substantially on the classifier \citep{Guyon2011,Evans2013}.
To provide model diversity, the study uses several classifiers with different potential capabilities: LDA, $5$-nn, na\"ive Bayes and SVM (see Section \ref{subsection:Classification}).
The classifier implementation details are described in Appendix C.

Many different classification problems are explored, including real and simulated data, described in Appendix B.
These problems are divided into three problem groups to clarify the results, see Section \ref{subsection:Assembly of Aggregate Results}.
The experimental study uses error rate for the loss function $L$ (see Section \ref{subsection:Classification}).
Further results are available for other loss functions (Brier Score and the H-measure) but are omitted for space.

The definition of $Q^c$ in Equation \ref{eq:eqc} conditions on certain sources of variation: the covariate location ${\bf x}$, the classifier $\theta$, the labelled dataset $D_S$, and the true class probability vector ${\bf p}$.
The experimental study explores several sources of variation: the AL algorithms, the classifier $\theta$, 
and the classification problem $({\bf X}, Y)$.

\subsection{Active Learning Methods}
\label{subsection:Active Learning Methods}

The experimental study evaluates many AL methods, to compare their performance across a range of classification problems.
These methods fall into three groups: RS as the natural benchmark of AL, standard AL methods from the literature, and algorithms estimating $Q^c$.
The second group consists of four standard AL methods: SE, QbcV, QbcA, and EfeLc (all described in Section \ref{subsection:Literature Review}).
The third group contains the two $Q^c$-estimation algorithms, simpleEQ and partitionEQ, defined in Section \ref{section:Algorithms to Estimate Example Quality} and abbreviated as SEQ and PEQ.

For the two Qbc methods, a committee of four classifiers is chosen for model diversity: logistic regression, $5$-nn, $21$-nn, and random forest.
Logistic regression is a parametric discriminative classifier described in \citet{Schein2007}; random forest is a non-parametric classifier described in \citet{Breiman2001}; $K$-nn is described in Section \ref{subsection:Classification}.
This committee is arbitrary, but diverse; the choices of committee size and constitution are open research problems.

Density weighting is sometimes recommended in the AL literature, see \citet{Olsson2009}.
However, the effects of density weighting are not theoretically understood. 
The experimental study also generated results from density weighting, omitted due to space, which left unaltered the primary conclusion that $Q^c$-estimation algorithms are competitive with standard methods from the literature.
The issue of density weighting is deferred to future work.

\subsection{Experimental AL Sandbox}

Iterated AL provides for the exploration of AL performance across the whole learning curve, see Section \ref{subsection:Active Learning} and \citet{Guyon2011,Evans2013}.
In this experimental study, the AL iteration continues until the entire pool has been labelled.
Each single experiment has a specific context: a classification problem, a base classifier, and a random seed.
Monte Carlo replication is applied to the experiments via the random seed, which is used to reshuffle the classification data.
The seed thereby affects the pool $X_P$ and the test data $D_E$, which are both drawn from the reshuffled classification data; ten seeds are used.

\begin{figure}
\centering
\includegraphics[scale=0.8]{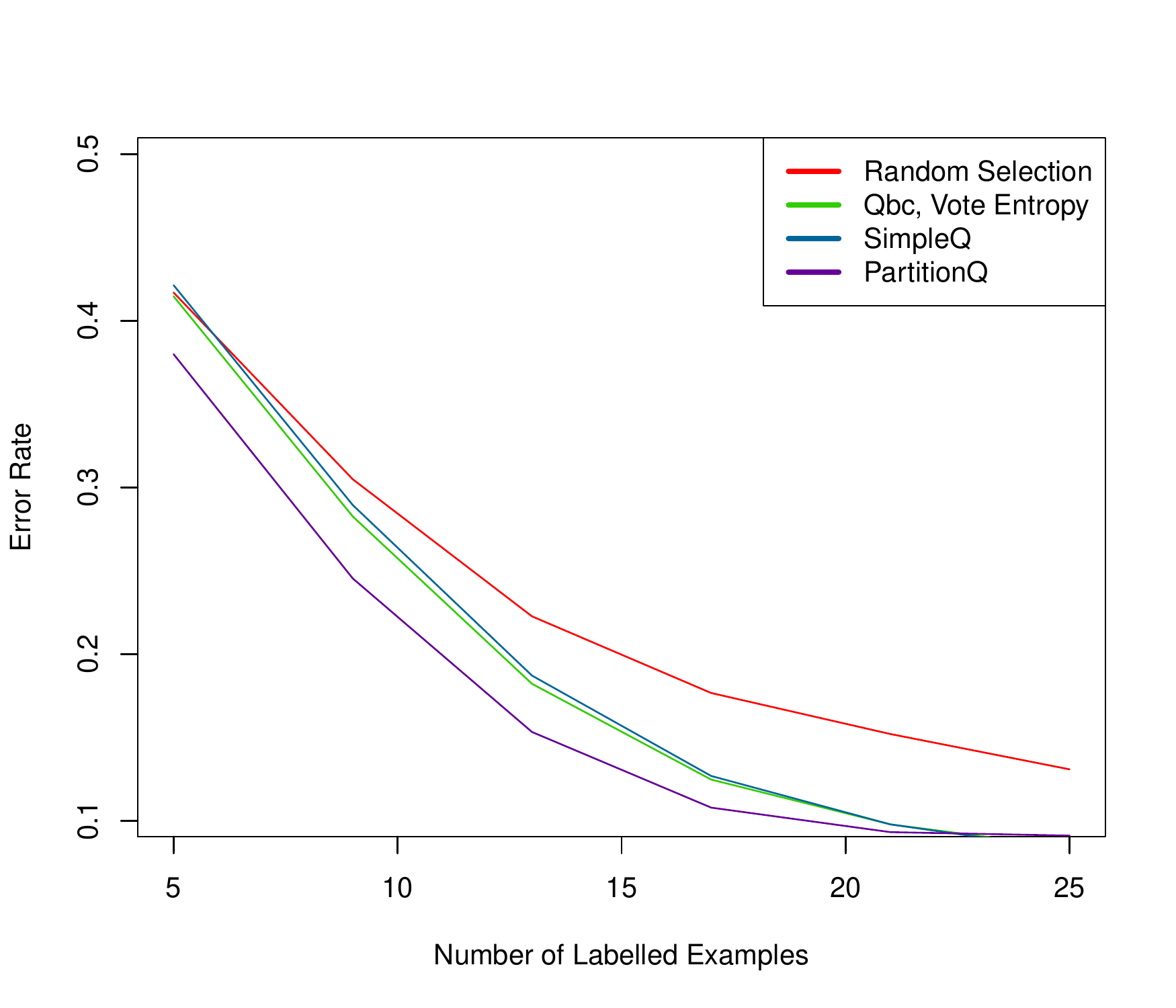}
\caption{Result for a single experiment of iterated AL. 
Each AL method performs multiple selection steps, generating a set of losses that define the learning curve. 
For clarity, a smoothed representation of the data is presented.
The early part of the learning curve is shown; beyond that, all learning curves tend to the same asymptote.
The classification problem is the Ripley four-Gaussian problem (see Figure \ref{fig:Qc rankings of Ripley to show similarity of rankings 1} and Appendix B), with the base classifier being $5$-nn.
}
\label{figure:Single result for simulated data 1}
\end{figure}

Given this experimental context, the experimental AL sandbox then evaluates the performance of all AL methods over a single dataset, using iterated AL.
Each AL method produces a learning curve that shows the overall profile of loss as the number of labelled examples increases.
To illustrate, Figure \ref{figure:Single result for simulated data 1} shows the learning curve for several AL methods.

\subsection{Assessing Performance}
\label{subsection:Assessing Performance}

The AL literature provides a selection of metrics to assess AL performance, such as AUA \citep{Guyon2011}, WI \citep{Evans2013} and label complexity \citep{Dasgupta2011}.
Experimental results suggest substantial agreement between the metrics, in their ranking of AL methods.
For that reason, and to avoid any arbitrary choice of one single metric, this study employs several metrics.

The experimental study evaluates four metrics: AUA, WI with two weighting functions (exponential with $\alpha=0.02$, and linear), and label complexity (with 
$\epsilon=5$).
The \emph{overall rank} is also calculated as the ranking of the mean ranks.
This yields five AL metrics: four primary metrics (label complexity, AUA, WI-linear, WI-exponential) and one aggregate metric (overall rank).

As discussed in Section \ref{subsection:Active Learning}, AL performance metrics assess the relative improvements in classifier performance, when comparing one AL method against another (or when comparing AL against RS).
Thus the real quantity of interest is the ranking of the AL methods. 

For a single experiment, there is a single classification problem and base classifier.
In such an experiment, all five metrics are evaluated for every AL method, so that each metric produces its own ranking of the AL methods.
Since there are seven AL methods (see Section \ref{subsection:Active Learning Methods}), the ranks fall between one and seven, with some ties.

The results show that the AL metrics substantially agree on AL method ranking (see Tables \ref{table:Single Pairing Result 1} and \ref{table:Problem Set Result 1}).
This agreement suggests that the ranking results are reasonably insensitive to the choice of AL metric.
In this experimental study, AL performance is assessed by overall rank. 

\subsection{Assembly of Aggregate Results}
\label{subsection:Assembly of Aggregate Results}

To address the variability of AL, multiple Monte Carlo draws are conducted for each dataset.
Thus for each experiment, the labelled, pool and test data are drawn from the population, as different independent subsamples.
This random subsampling addresses two primary sources of variation, namely the initially labelled data and the unlabelled pool.

The experimental study examines many Monte Carlo draws, classification problems in groups, and classifiers.
The aggregate results are calculated by averaging, first over Monte Carlo replicates, and then over classification problems.

\begin{table}[h!b!p!] 
\caption{Results for a single pairing of classifier and problem. The base classifier is $5$-nn. The classification problem is Credit-95. The six methods shown are selected as the best by overall rank.}
\label{table:Single Pairing Result 1}
\centering
\begin{tabular}{|c|c|c|c|c|c|c|}
\hline
\multicolumn{7}{c}{Classifier $5$-nn, and Credit-95 Problem}\\
\hline
\hline
AL Performance Metric & Method 1 & Met. 2 & Met. 3 & Met. 4 & Met. 5 & Met. 6 \\ 
\hline
\hline
 & SE & QbcV & \textbf{SEQ} & \textbf{PEQ} & RS & QbcA \\
\hline
\hline
\makecell{Label Complexity} & 4 & 2 & 1 & 6 & 2 & 5\\
\hline
\makecell{AUA} & 1 & 2 & 3 & 4 & 5 & 6\\
\hline
\makecell{WI-Linear} & 1 & 2 & 6 & 3 & 5 & 4\\
\hline
\makecell{WI-Exponential} & 1 & 2 & 4 & 3 & 6 & 5\\
\hline
\makecell{Overall Rank} & 1 & 2 & 3 & 4 & 5 & 6\\ 
\hline
\end{tabular}
\end{table}

\begin{table}[h!b!p!] 
\caption{Results for a single classifier and a group of problems. 
The base classifier is $5$-nn. 
The classification problem group is the large problem group (see Appendix B).
The six methods shown are selected as the best by overall rank.}
\label{table:Problem Set Result 1}
\centering
\begin{tabular}{|c|c|c|c|c|c|c|}
\hline
\multicolumn{7}{c}{Classifier $5$-nn, and large data problem group}\\
\hline
\hline
AL Performance Metric & Method 1 & Met. 2 & Met. 3 & Met. 4 & Met. 5 & Met. 6 \\
\hline
\hline
 & SE & \textbf{PEQ} & \textbf{SEQ} & QbcV & QbcA & RS \\
\hline
\hline
\makecell{Label Complexity} & 7 & 4 & 2 & 3 & 5 & 1\\
\hline
\makecell{AUA} & 1 & 3 & 2 & 4 & 5 & 6\\
\hline
\makecell{WI-Linear} & 1 & 2 & 5 & 4 & 3 & 6\\
\hline
\makecell{WI-Exponential} & 1 & 2 & 3 & 4 & 5 & 6\\ 
\hline
\makecell{Overall Rank} & 1 & 2 & 3 & 4 & 5 & 6\\
\hline
\end{tabular}
\end{table}

First the results over the Monte Carlo replicates are averaged, to produce the aggregate result for a given pairing of classifier and problem.
Table \ref{table:Single Pairing Result 1} is shown just to illustrate one such result, where six AL methods are chosen as the best by overall rank.
Second, further averaging over the problem within a group yields the aggregate result for a pairing of classifier and problem group; this second averaging is shown in Table \ref{table:Problem Set Result 1}.
Finally the results are assembled for all three problem groups, to form the aggregate results for a single classifier, shown in Table \ref{table:LDA Result Original 1}.
For each group of problems, six methods are judged as the best by overall rank. 

\begin{table}[h!b!p!] 
\caption{
Results for base classifier LDA over three groups of problems.
The six methods shown are selected as the best by overall rank.}
\label{table:LDA Result Original 1}
\centering
\begin{tabular}{|c|c|c|c|c|c|c|}
\hline
\multicolumn{7}{c}{Classifier LDA}\\
\hline
\hline
 & Method 1 & Met. 2 & Met. 3 & Met. 4 & Met. 5 & Met. 6 \\
\hline
\hline
\makecell{Small Problems} & QbcA & QbcV & SE & \textbf{PEQ} & \textbf{SEQ} & RS \\
\hline
\makecell{Large Problems} & \textbf{PEQ} & QbcV & SE & RS & \textbf{SEQ} & QbcA \\
\hline
\makecell{Theoretical Problems} & QbcV & SE & \textbf{PEQ} & \textbf{SEQ} & QbcA & RS \\
\hline
\end{tabular}
\end{table}

\subsection{Results}
\label{subsection:Primary Results}

\begin{table}[h!b!p!] 
\caption{
Results for base classifier $5$-nn over three groups of problems.
The six methods shown are selected as the best by overall rank. }
\label{table:KNN Result Original 1}
\centering
\begin{tabular}{|c|c|c|c|c|c|c|}
\hline
\multicolumn{7}{c}{Classifier $5$-nn}\\
\hline
\hline
 & Method 1 & Met. 2 & Met. 3 & Met. 4 & Met. 5 & Met. 6 \\
\hline
\hline
\makecell{Small Problems} & SE & RS & QbcV & QbcA & \textbf{PEQ} & \textbf{SEQ} \\
\hline
\makecell{Large Problems} & SE & \textbf{PEQ} & \textbf{SEQ} & QbcA & QbcV & RS \\ 
\hline
\makecell{Theoretical Problems} & \textbf{PEQ} & QbcV & SE & RS & \textbf{SEQ} & QbcA \\
\hline
\end{tabular}
\end{table}

\begin{table}[h!b!p!] 
\caption{
Results for base classifier na\"ive Bayes over three groups of problems.
The six methods shown are selected as the best by overall rank.}
\label{table:Naive Bayes Result Original 1}
\centering
\begin{tabular}{|c|c|c|c|c|c|c|}
\hline
\multicolumn{7}{c}{Classifier na\"ive Bayes}\\
\hline
\hline
 & Method 1 & Met. 2 & Met. 3 & Met. 4 & Met. 5 & Met. 6 \\
\hline
\hline
\makecell{Small Problems} & RS & QbcV & SE & \textbf{SEQ} & \textbf{PEQ} & QbcA \\
\hline
\makecell{Large Problems} & QbcV & SE & \textbf{SEQ} & \textbf{PEQ} & RS & QbcA \\
\hline
\makecell{Theoretical Problems} & QbcV & SE & QbcA & RS & \textbf{PEQ} & \textbf{SEQ} \\
\hline
\end{tabular}
\end{table}

\begin{table}[h!b!p!] 
\caption{
Results for base classifier SVM over three groups of problems.
The six methods shown are selected as the best by overall rank.}
\label{table:SVM Result Original 1}
\centering
\begin{tabular}{|c|c|c|c|c|c|c|}
\hline
\multicolumn{7}{c}{Classifier SVM}\\
\hline
\hline
 & Method 1 & Met. 2 & Met. 3 & Met. 4 & Met. 5 & Met. 6 \\
\hline
\hline
\makecell{Small Problems} & QbcV & QbcA & \textbf{PEQ} & RS & SE & \textbf{SEQ} \\
\hline
\makecell{Large Problems} & QbcV & QbcA & \textbf{PEQ} & SE & \textbf{SEQ} & RS \\
\hline
\makecell{Theoretical Problems} & QbcV & RS & \textbf{PEQ} & QbcA & \textbf{SEQ} & SE \\
\hline
\end{tabular}
\end{table}



The results for LDA, $K$-nn, na\"ive Bayes and SVM are shown in Tables \ref{table:LDA Result Original 1}, \ref{table:KNN Result Original 1}, \ref{table:Naive Bayes Result Original 1} and \ref{table:SVM Result Original 1} respectively.
These results are the primary results of the of the experimental study, covering the four classifiers, all the problems in three groups, and multiple Monte Carlo replicates.

For all four classifiers, the $Q^c$-motivated algorithms perform effectively in comparison to the standard AL methods from the literature.
This conclusion holds true over different classifiers and different classification problems.
This reinforces the advantages of the explicit theoretical target $Q^c$. 

Considering the AL methods from the literature, QBC and SE consistently perform well.
For QBC, vote entropy (QbcV) mostly outperforms average Kullback-Liebler divergence (QbcA).
EfeLc performs somewhat less well, perhaps because of the way it approximates loss using the unlabelled pool (see Section \ref{subsection:Literature Review}).

Comparing the $Q^c$-estimation algorithms, the algorithm partitionEQ outperforms the algorithm simpleEQ, in all cases except two.
The fact that the data partitioning approach of partitionEQ outperforms the naive reuse approach of simpleEQ may have implications for $Q^c$-estimation algorithms in general.
This suggests that more sophisticated forms of data partitioning and resampling may improve $Q^c$-estimation algorithms.
Such algorithms are the subject of future work.

\section{Conclusion}

Example quality is a new theoretical approach to AL which defines optimal behaviour via loss.
This provides the abstract target $Q^c$, which is defined for any classifier and for any loss function.

Exploring this abstract definition of optimal AL behaviour generates new insights into AL.
For example, the optimal selection is examined and compared to known AL methods, revealing suboptimal choices.
Further, the dependence of the target $Q^c$ on the labelled data motivates the definition and exploration of optimal behaviour independent of the labelled data.

While the approach is primarily theoretical, example quality also addresses AL applications, by defining $Q^c$ as a target for algorithms to estimate.
A comprehensive experimental study compares the performance of $Q^c$-estimation algorithms alongside several standard AL methods.
The results demonstrate that the $Q^c$ algorithms are competitive across a range of classifiers and problems.
Thus example quality provides insights into AL, and motivates methods to address example selection in practice.

This framework lays the ground for future theoretical examination of other aspects of AL, particularly density weighting.
This work defines two straightforward algorithms to estimate $Q^c$; more sophisticated estimators are the subject of future research.




\acks{The work of Lewis P. G. Evans is supported by a doctoral training award from the EPSRC.}


\appendix

\section*{Appendix A.}
\label{section:Appendix A.}

Section \ref{subsection:Theoretical Example to Illustrate Example Quality} describes a theoretical classification problem and calculates $Q^c$ as an explicit function of $x$ for error loss $L_e$.
This Appendix presents an analytic calculation of the target $Q^m$ for error loss $L_e$, and a brief exploration of the behaviour of $Q^m$.
Finally the calculation of error loss $L_e$ is given.
Further results are available but omitted for space, such as the targets $Q^c$ and $Q^m$ for other loss functions (quadratic loss and Brier Score), see \citet{Hand1997}.


\subsection*{Analytic Calculation of $Q^m$ with Error Loss}

This is the analytic calculation of $Q^m$ under error loss $L_e$. 
In this Appendix, $n_s$ is abbreviated to $n$. 

To recap from Section \ref{subsection:Theoretical Example to Illustrate Example Quality}, $(X|Y=c_1) \sim \textrm{N} (-1, 1)$, $(X|Y=c_2) \sim \textrm{N} (1, 1)$ and $\hat{t} \sim \textrm{N} (0, \frac{1}{n})$.

The true densities $q_j(x)$ and estimated densities $\hat{q}_j(x)$ of the Gaussian class-conditional densities $(X|Y=c_j)$ are given by 
\begin{equation*}
q_j(x) = \frac{1}{\sqrt{2\pi}} \exp{\frac{-1}{2}(x - \mu_j)^2} \, \, \, \textrm{ ; } \, \, \, \hat{q}_j(x) = \frac{1}{\sqrt{2\pi}} \exp{\frac{-1}{2}(x - \hat{\mu}_j)^2}.
\end{equation*}
The true class probabilities $p_j(x)$ and estimated class probabilities $\hat{p}_j(x)$ of the Gaussian class-conditional densities $(X|Y=c_j)$ are given by 
\begin{equation*}
p_j(x) = \frac{q_j(x)}{q_1(x) + q_2(x)} \, \, \, \textrm{ ; } \, \, \, \hat{p}_j(x) = \frac{\hat{q}_j(x)}{\hat{q}_1(x) + \hat{q}_2(x)}.
\end{equation*}

Further, $\hat{t}^{\prime}_j = \hat{t} + \frac{z}{2}(x - \hat{\mu}_j) = (\hat{t} + \frac{z x}{2}) - (\frac{z \hat{\mu}_j}{2})$.
For class $c_1$, $\hat{t}^{\prime}_1 = (\hat{t} + \frac{z x}{2}) - (\frac{z \hat{\mu}_1}{2})$; of those final two terms, the sampling distributions are $\textrm{N}(\frac{z x}{2}, \frac{1}{n})$ and $\textrm{N}(\frac{-z}{2}, \frac{2 z^2}{4n})$ respectively.
Hence the sampling distribution of $\hat{t}^{\prime}_1$ is $\hat{t}^{\prime}_1 \sim \textrm{N}(\frac{z}{2}(x+1), \frac{2+z^2}{2n})$.

Whereas for class $c_2$, $\hat{t}^{\prime}_2 = (\hat{t} + \frac{z x}{2}) - (\frac{z \hat{\mu}_2}{2})$; of those final two terms, the sampling distributions are $\textrm{N}(\frac{z x}{2}, \frac{1}{n})$ and $\textrm{N}(\frac{z}{2}, \frac{z^2}{4 n})$ respectively.
Hence the sampling distribution of $\hat{t}^{\prime}_2$ is $\hat{t}^{\prime}_2 \sim \textrm{N}(\frac{z}{2}(x-1), \frac{2+z^2}{2n})$.

First consider the expectation of Term F.
Equation \ref{eq:le_equation_1} gives 
\begin{equation*}
\begin{split}
L_e(\hat{\mu}_1, \hat{\mu}_2) = \frac{1}{2} \{ 1 - F_1(\hat{t}) + F_2(\hat{t}) + \mathbbm{1}(\hat{\mu}_1 > \hat{\mu}_2) [2 F_1(\hat{t}) - 2 F_2(\hat{t})] \}.
\end{split}
\end{equation*} 

For $F_1$, $F_1(\hat{t}) = P(q_1(x) < \hat{t}) = P(q_1(x) - \hat{t} < 0) = F_{Z_{10}}(0)$ for $Z_{10} = q_1(x) - \hat{t}$.
The sampling distributions are given by $q_1 \sim \textrm{N}(-1,1)$ and $\hat{t} \sim \textrm{N}(0, \frac{1}{n})$.
Hence $Z_{10} \sim \textrm{N}(-1, \frac{n+1}{n})$.

For $F_2$, $F_2(\hat{t}) = P(q_2(x) < \hat{t}) = P(q_2(x) - \hat{t} < 0) = F_{Z_{11}}(0)$ for $Z_{11} = q_2(x) - \hat{t}$.
The sampling distributions are given by $q_2 \sim \textrm{N}(+1,1)$ and $\hat{t} \sim \textrm{N}(0, \frac{1}{n})$.
Hence $Z_{11} \sim \textrm{N}(+1, \frac{n+1}{n})$.

Further, $P(\mathbbm{1}(\hat{\mu}_1 > \hat{\mu}_2) = P(\hat{\mu}_1 > \hat{\mu}_2) = P(\hat{\mu}_1 - \hat{\mu}_2 > 0) = 1 - F_{Z_{12}}(0)$, for $Z_{12} = \hat{\mu}_1 - \hat{\mu}_2$, hence $Z_{12} \sim \textrm{N}(-2, \frac{4}{n})$.
Combining these sampling distributions with the definition of Term F as $L_e(\hat{\mu}_1, \hat{\mu}_2)$ 
yields 
$$
E[\textrm{Term F}] = \frac{1}{2} \{ 1 - F_{Z_{10}}(0) + F_{Z_{11}}(0) + [1 - F_{Z_{12}}(0)] [2 F_{Z_{10}}(0) - 2 F_{Z_{11}}(0)] \}.
$$

Turning to Term J which is given by
\begin{equation*}
\begin{split}
\textrm{Term J} = \frac{p_1}{2} \{ 1 - F_1(\hat{t}^{\prime}_1) + F_2(\hat{t}^{\prime}_1) + \mathbbm{1}(\hat{\mu}^{\prime}_1 > \hat{\mu}_2) [2 F_1(\hat{t}^{\prime}_1) - 2 F_2(\hat{t}^{\prime}_1)] \} \\
+ \frac{p_2}{2} \{ 1 - F_1(\hat{t}^{\prime}_2) + F_2(\hat{t}^{\prime}_2) + \mathbbm{1}(\hat{\mu}_1 > \hat{\mu}^{\prime}_2) [2 F_1(\hat{t}^{\prime}_2) - 2 F_2(\hat{t}^{\prime}_2)] \}.
\end{split}
\end{equation*} 

The sampling distributions of $\hat{t}^{\prime}_1$ and $\hat{t}^{\prime}_2$ are given above.
$F_1(\hat{t}^{\prime}_1) = P(q_1(x) < \hat{t}^{\prime}_1) = P(q_1(x) - \hat{t}^{\prime}_1 < 0) = F_{Z_3}(0)$ for $Z_{3} = q_1(x) - \hat{t}^{\prime}_1$.
Hence $Z_{3} \sim \textrm{N}(-1 - \frac{z}{2}(x+1), 1 + \frac{2+z^2}{2n})$.
$F_2(\hat{t}^{\prime}_1) = P(q_2(x) < \hat{t}^{\prime}_1) = P(q_2(x) - \hat{t}^{\prime}_1 < 0) = F_{Z_6}(0)$ for $Z_{6} = q_2(x) - \hat{t}^{\prime}_1$.
Hence $Z_{6} \sim \textrm{N}(1 - \frac{z}{2}(x+1), 1 + \frac{2+z^2}{2n})$.
Further, $P(\mathbbm{1}(\hat{\mu}^{\prime}_1 > \hat{\mu}_2) = P(\hat{\mu}^{\prime}_1 > \hat{\mu}_2) = P(\hat{\mu}^{\prime}_1 - \hat{\mu}_2 > 0) = 1 - F_{Z_{13}}(0)$, for $Z_{13} = \hat{\mu}^{\prime}_1 - \hat{\mu}_2$, hence $Z_{13} \sim \textrm{N}(z + zx - 2, \frac{2}{n}(z^2 - 2z + 2))$.

Further, $F_1(\hat{t}^{\prime}_2) = P(q_1(x) < \hat{t}^{\prime}_2) = P(q_1(x) - \hat{t}^{\prime}_2 < 0) = F_{Z_8}(0)$ for $Z_{8} = q_1(x) - \hat{t}^{\prime}_2$.
Hence $Z_{8} \sim \textrm{N}(-1 - \frac{z}{2}(x-1), 1 + \frac{2+z^2}{2n})$.
$F_2(\hat{t}^{\prime}_2) = P(q_2(x) < \hat{t}^{\prime}_2) = P(q_2(x) - \hat{t}^{\prime}_2 < 0) = F_{Z_9}(0)$ for $Z_{9} = q_2(x) - \hat{t}^{\prime}_2$.
Hence $Z_{9} \sim \textrm{N}(1 - \frac{z}{2}(x-1), 1 + \frac{2+z^2}{2n})$.
Further, $P(\mathbbm{1}(\hat{\mu}_1 > \hat{\mu}^{\prime}_2) = P(\hat{\mu}_1 > \hat{\mu}^{\prime}_2) = P(\hat{\mu}_1 - \hat{\mu}^{\prime}_2 > 0) = 1 - F_{Z_{16}}(0)$, for $Z_{16} = \hat{\mu}_1 - \hat{\mu}^{\prime}_2$, hence $Z_{16} \sim \textrm{N}(z - zx - 2, \frac{2}{n}(z^2 - 2z + 2))$.

Using these sampling distributions with the definition of Term J above yields
\begin{align*} 
Q^m (x, n) & = E [\textrm{Term F}] - E [\textrm{Term J}] \\
& = \frac{1}{2} \{ 1 - F_{Z_{10}}(0) + F_{Z_{11}}(0) + [1 - F_{Z_{12}}(0)] [2 F_{Z_{10}}(0) - 2 F_{Z_{11}}(0)] \} \nonumber \\
& - \frac{p_1}{2} \{ 1 - F_{Z_{3}}(0) + F_{Z_{6}}(0) + [1 - F_{Z_{13}}(0)] [2 F_{Z_{3}}(0) - 2 F_{Z_{6}}(0)] \} \nonumber \\
& - \frac{p_2}{2} \{ 1 - F_{Z_{8}}(0) + F_{Z_{9}}(0) + [1 - F_{Z_{16}}(0)] [2 F_{Z_{8}}(0) - 2 F_{Z_{9}}(0)] \} \nonumber,
\end{align*}
where the distributions of the $Z_i$ variables are given above.

\begin{figure}
        \centering
        \begin{subfigure}[b]{0.45\textwidth}
                \centering
               \includegraphics[width=\textwidth]{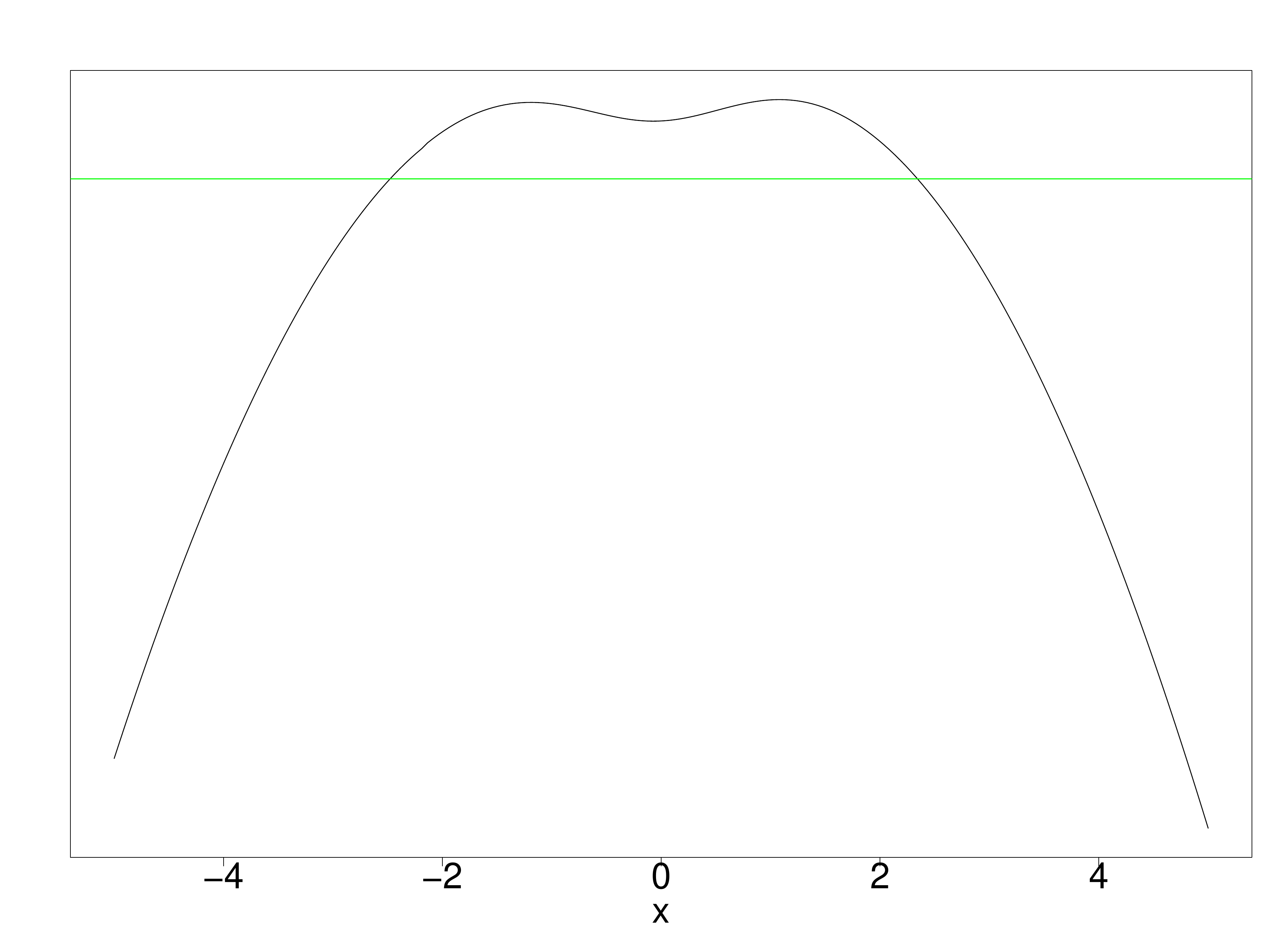}
        \end{subfigure}


        \caption{Illustration of the target $Q^m$ as a function of $x$.
The green line indicates $Q^m(x)= 0$ (zero improvement), with $n_s$ being 18.
The substantial set of central $x$ values where $Q^m$ is positive shows that many choices of the example $x$ will improve the classifier.
}
	\label{fig:Qm for two loss functions}
\end{figure}

Figure \ref{fig:Qm for two loss functions} illustrates $Q^m$ as a function of $x$.
This expression for $Q^m$ is a complicated non-linear function of $x$, due to the non-linearity of $p_1(x)$, $p_2(x)$ and $F_{Z_i}(0)$.
The substantial set of central $x$ values where $Q^m$ is positive shows that many choices of $x$ will improve the classifier.

\subsection*{Analytic Calculation of Error Loss}

First the classifier decision rule $r_1(x)$ given in Section \ref{subsubsection:Calculation of Qc} can be expressed as 
\begin{displaymath}
   r_2(x) = \left\{
     \begin{array}{lr}
       \hat{y} = c_1 & : x < \hat{t}, \\
       \hat{y} = c_2 & : x > \hat{t}, \\
     \end{array}
   \right.
   r_3(x) = \left\{
     \begin{array}{lr}
       \hat{y} = c_1 & : x > \hat{t}, \\
       \hat{y} = c_2 & : x < \hat{t}, \\
     \end{array}
   \right.
   r_1(x) = \left\{
     \begin{array}{lr}
       \hat{y} = r_2(x) & : \hat{\mu}_1 < \hat{\mu}_2, \\
       \hat{y} = r_3(x) & : \hat{\mu}_1 > \hat{\mu}_2, \\
     \end{array}
   \right.
\end{displaymath}

The probability of an error for a single covariate $x$ is denoted $p(e|x) = p(y=c_1, \hat{y}=c_2 | x) + p(y=c_2, \hat{y}=c_1 | x)$.
Given $x$, the probabilities of $y$ and $\hat{y}$ are independent. 
Thus for decision rule $r_2$, the loss $L_e$ is given by 

\begin{align*}
L_e(\boldsymbol{\hat{\theta}}) 
& = E_X [p(e|x)] = \int p(y=c_1, \hat{y}=c_2 | x) p(x) dx + \int p(y=c_2, \hat{y}=c_1 | x) p(x) dx \\
& = \int p_1(x) 1(x > \hat{t}) p(x) dx + \int p_2(x) 1(x < \hat{t}) p(x) dx \\ 
& = \int \frac{1}{2} q_1(x) 1(x > \hat{t}) dx + \int \frac{1}{2} q_2(x) 1(x < \hat{t}) dx
 = \frac{1}{2} [ \int^{\infty}_{\hat{t}} q_1(x) dx + \int^{\hat{t}}_{-\infty} q_2(x) dx ] \\
& = \frac{1}{2} [1 - F_1(\hat{t}) + F_2(\hat{t}) ].
\end{align*}

Whereas for decision rule $r_3$, the loss $L_e$ is given by $L_e(\boldsymbol{\hat{\theta}}) = \frac{1}{2} [1 + F_1(\hat{t})  - F_2(\hat{t}) ]$.
Hence the loss $L_e$ in general is given by
\begin{equation*} 
\begin{split}
L_e(\boldsymbol{\hat{\theta}}) = \frac{1}{2} \{ 1 - F_1(\hat{t}) + F_2(\hat{t}) + \mathbbm{1}(\hat{\mu}_1 > \hat{\mu}_2) [2 F_1(\hat{t}) - 2 F_2(\hat{t})] \}.
\end{split}
\end{equation*}

\section*{Appendix B.}

A diverse set of classification problems is chosen to explore AL performance.
The classification problems fall into two sets: real problems and theoretical problems.

First the real data classification problems are shown in Tables \ref{table:Small Real Data 1} and \ref{table:Large Real Data 1}.
The real data problems are split into two groups, one for smaller problems of fewer examples, and another of larger problems.
The class prior is shown, since the experimental study uses error rate as loss.
The sources for this data include UCI \citep{Bache2013}, \citet{Guyon2011}, \citet{Anagnostopoulos2012} and \citet{Adams2010}.

The intention here is to provide a wide variety in terms of problem properties: covariate dimension $d$, number of classes $k$, the class prior $\boldsymbol{\pi}$, and the underlying distribution.
The number and variety of problems suggests that the results in Section \ref{section:Experiments and Results} have low sensitivity to the presence or absence of one or two specific problems.

\begin{table}[h!b!p!] 
\caption{Real Data Classification Problems, Smaller}
\label{table:Small Real Data 1}
\centering
\begin{tabular}{|c|c|c|c|c|}
\hline
Name & Dim. $d$ & Classes $k$ & Cases $n$ & Class Prior $\boldsymbol{\pi}$ \\
\hline
\hline
\makecell{Abalone} & 8 & 3 & 4177 & (0.31, 0.32, 0.37)\\
\hline
\makecell{Australian} & 14 & 2 & 690 & (0.44, 0.56)\\
\hline
\makecell{Balance} & 4 & 3 & 625 & (0.08, 0.46, 0.46)\\
\hline
\makecell{Breast Cancer Wdbc} & 4 & 3 & 569 & (0.37, 0.63)\\
\hline
\makecell{German Credit} & 24 & 2 & 1000 & (0.3, 0.7)\\
\hline
\makecell{Glass} & 10 & 6 & 214 & (0.33,0.36,0.08,0.06,0.04,0.14)\\
\hline
\makecell{Heart-Statlog} & 13 & 2 & 270 & (0.65, 0.44)\\
\hline
\makecell{Ionosphere} & 34 & 2 & 351 & (0.64, 0.36)\\
\hline
\makecell{Monks-1} & 6 & 2 & 432 & (0.5, 0.5)\\
\hline
\makecell{Monks-2} & 6 & 2 & 432 & (0.5, 0.5)\\
\hline
\makecell{Monks-3} & 6 & 2 & 432 & (0.5, 0.5)\\
\hline
\makecell{Pima Diabetes} & 8 & 2 & 768 & (0.35, 0.65)\\
\hline
\makecell{Sonar} & 60 & 2 & 208 & (0.47, 0.53)\\
\hline
\makecell{Wine} & 13 & 3 & 178 & (0.33, 0.4, 0.27)\\
\hline
\end{tabular}
\end{table}

\begin{table}[h!b!p!] 
\caption{Real Data Classification Problems, Larger}
\label{table:Large Real Data 1}
\centering
\begin{tabular}{|c|c|c|c|c|}
\hline
Name & Dim. $d$ & Classes $k$ & Cases $n$ & Class Prior $\boldsymbol{\pi}$ \\
\hline
\hline
\makecell{Fraud} & 20 & 2 & 5999 & (0.167, 0.833)\\
\hline
\makecell{Electricity Prices} & 6 & 2 & 27552 & (0.585, 0.415)\\ 
\hline
\makecell{Colon} & 16 & 2 & 17076 & (0.406, 0.594)\\ 
\hline
\makecell{Credit 93} & 29 & 2 & 4406 & (0.007, 0.993)\\ 
\hline
\makecell{Credit 94} & 29 & 2 & 8493 & (0.091, 0.909)\\ 
\hline
\makecell{Credit 95} & 29 & 2 & 21076 & (0.117, 0.883)\\ 
\hline
\makecell{Credit 96} & 29 & 2 & 24396 & (0.111, 0.889)\\ 
\hline
\makecell{Credit 97} & 29 & 2& 21191 & (0.043, 0.957)\\ 
\hline
\makecell{AL Challenge, Alex} & 11 & 2 & 10000 & (0.270, 0.730)\\ 
\hline
\makecell{AL Challenge, Sina} & 92 & 2 & 20722 & (0.062, 0.938)\\ 
\hline
\end{tabular}
\end{table}

Second the theoretical classification problems are illustrated in Figure \ref{fig:Contour graphs to show the tasks}.
These theoretical problems are generated by sampling from known probability distributions.
The class-conditional distributions $({\bf X} | y = c_j)_1^k$ are either Gaussians or mixtures of Gaussians. 
This set of problems presents a variety of decision boundaries to the classifier.
All have balanced uniform priors, and the Bayes Error Rates are approximately 0.1.


\def \Sim_Problem_Textwidth_Multiplier {0.19}

\begin{figure}
        \centering
        \begin{subfigure}[b]{\Sim_Problem_Textwidth_Multiplier \textwidth}
                \centering
                \includegraphics[width=\textwidth]{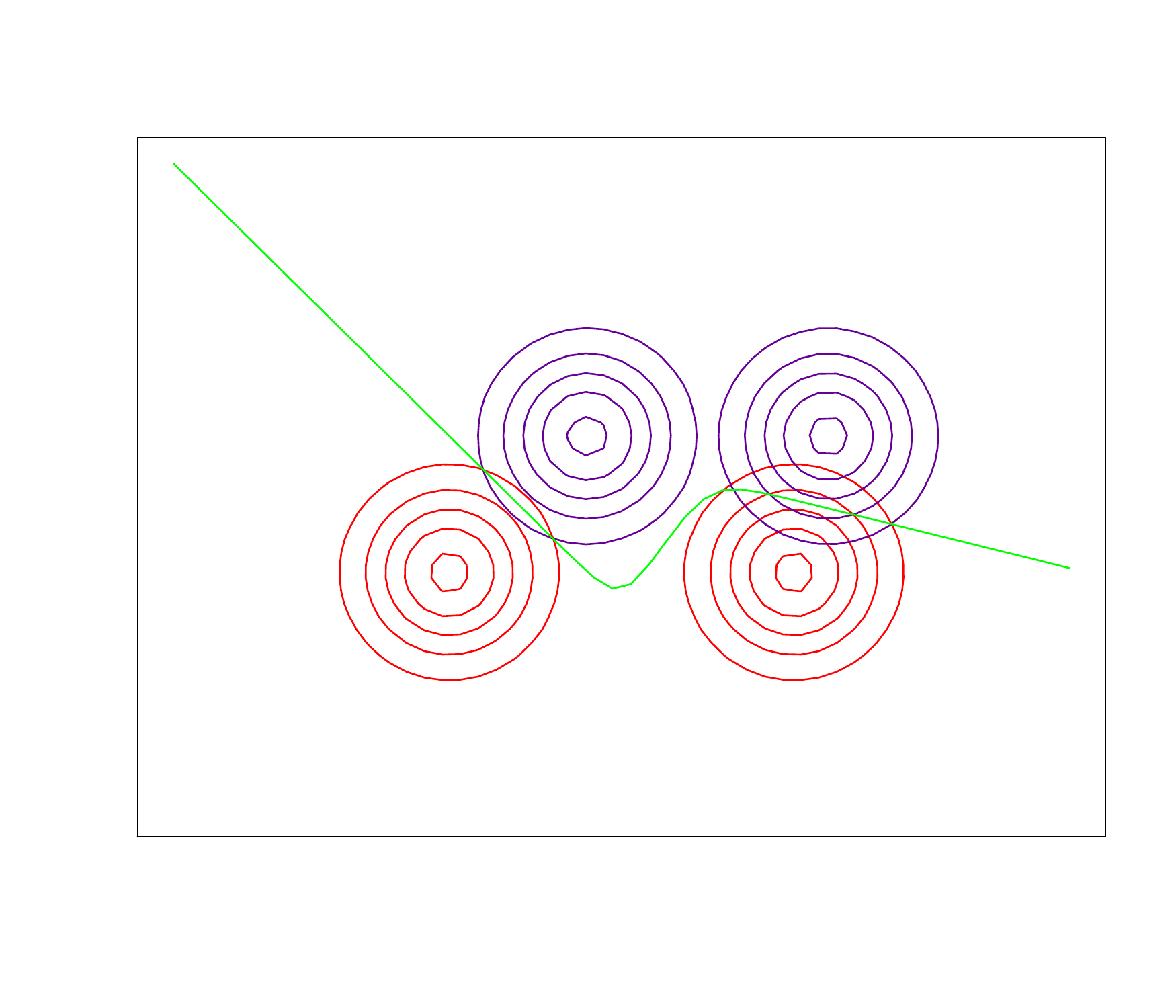} 
                \caption{Ripley \\ four-Gaussian \\ see \citep{Ripley1996}}
        \end{subfigure}
        \begin{subfigure}[b]{\Sim_Problem_Textwidth_Multiplier \textwidth}
                \centering
                \includegraphics[width=\textwidth]{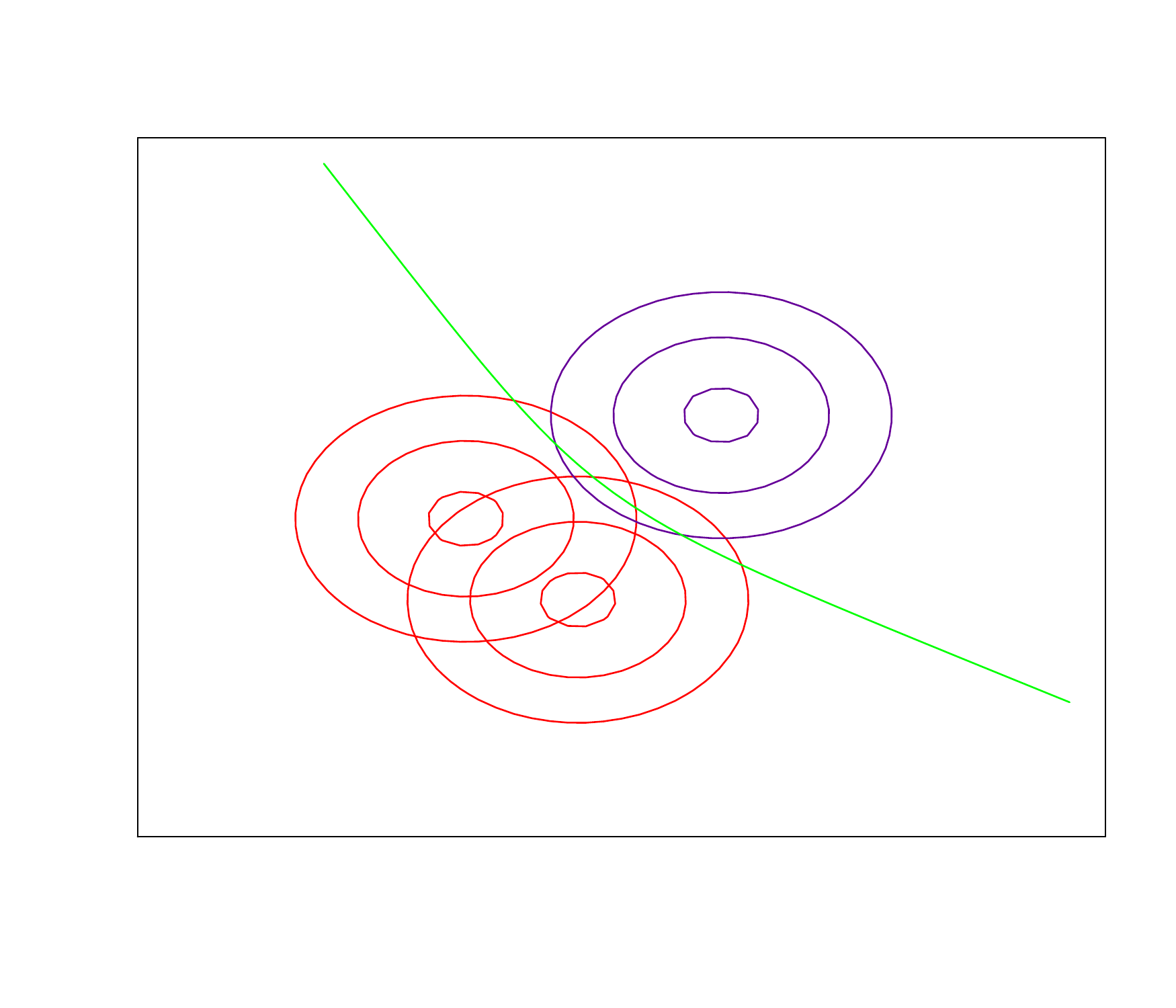} 
                \caption{Gaussian \\ Quadratic \\ boundary}
        \end{subfigure}%
        \begin{subfigure}[b]{\Sim_Problem_Textwidth_Multiplier \textwidth}
                \centering
                \includegraphics[width=\textwidth]{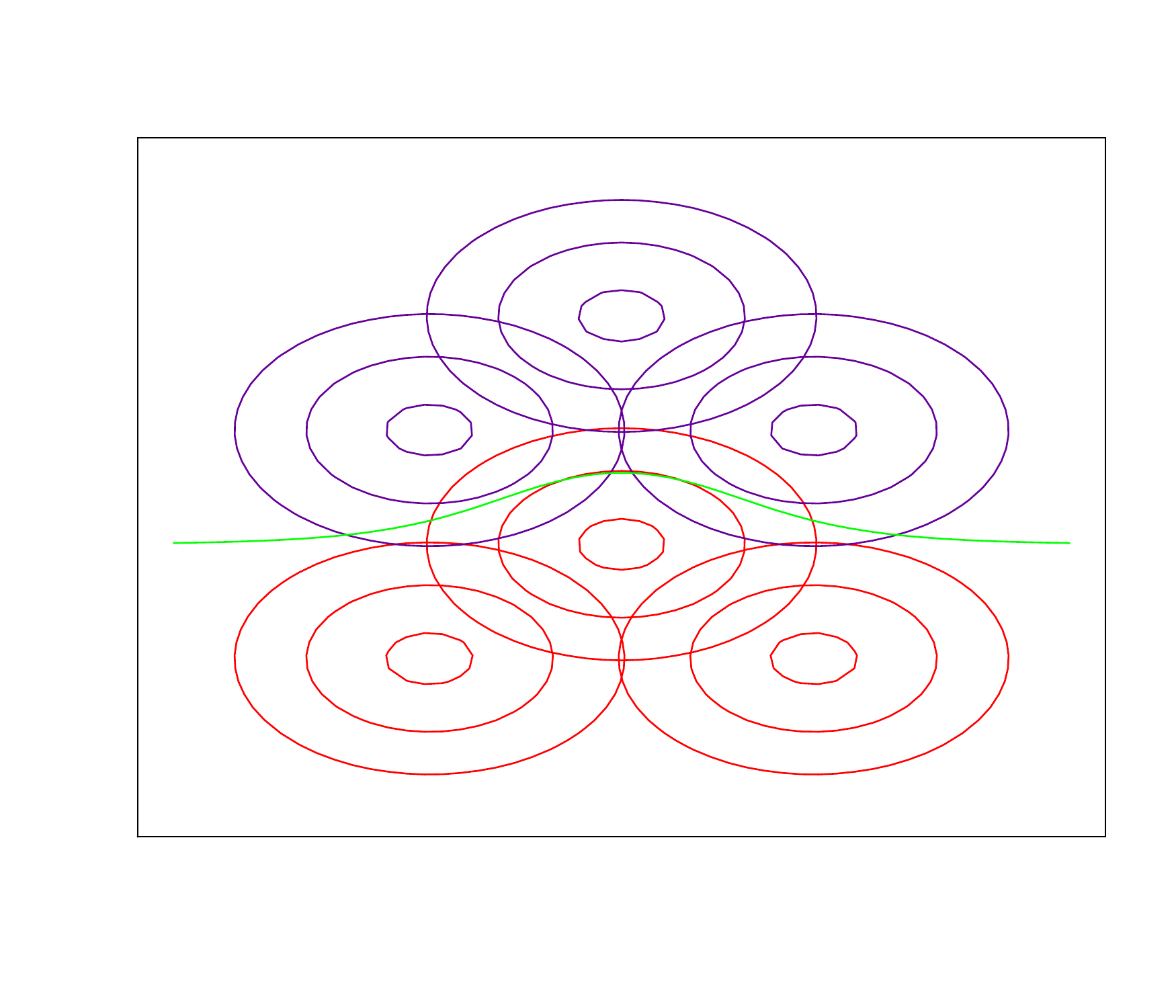} 
                \caption{Triangles of \\ three Gaussians\\}
        \end{subfigure}
        \begin{subfigure}[b]{\Sim_Problem_Textwidth_Multiplier \textwidth}
                \centering
                \includegraphics[width=\textwidth]{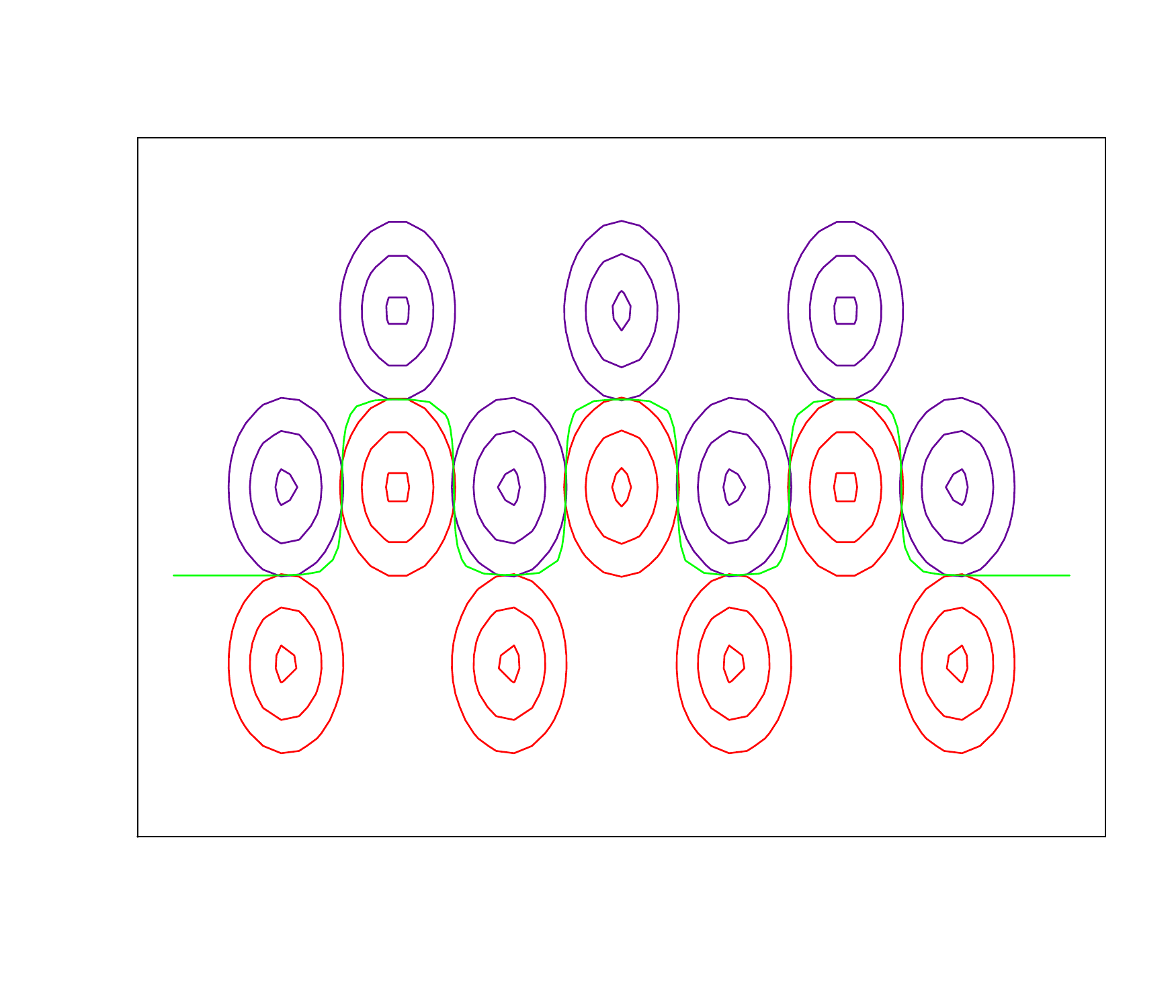} 
                \caption{Gaussian sets\\ oscillating-\\boundary}
        \end{subfigure}
        \centering
        \begin{subfigure}[b]{\Sim_Problem_Textwidth_Multiplier \textwidth}
                \centering
                \includegraphics[width=\textwidth]{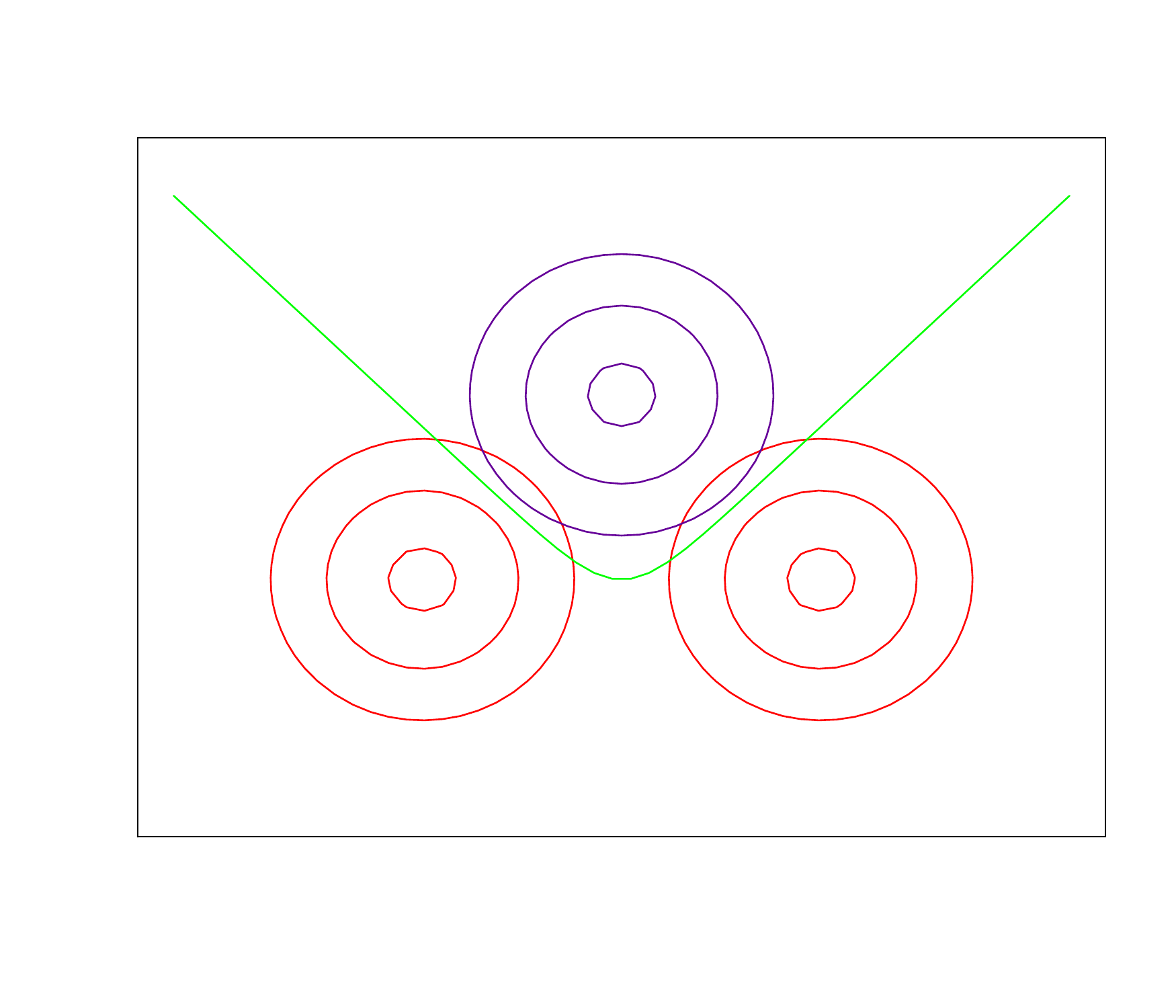} 
                \caption{Gaussian \\ sharply non-linear\\boundary} 
        \end{subfigure}
     
        \caption{Density contour plots showing the theoretical classification problems. 
The class-conditional distributions $({\bf X} | y = c_j)_1^k$ are shown in red for class 1 and blue for class 2.
These class-conditional distributions $({\bf X} | y = c_j)_1^k$ are either Gaussians or mixtures of Gaussians.
The decision boundary is shown in green.
}
	  \label{fig:Contour graphs to show the tasks}
\end{figure}

\section*{Appendix C.}

Here the classifier implementation details are described.
For LDA, the standard R implementation is used.
For $5$-nn, the R implementation from package kknn is used.
This implementation applies covariate scaling: each covariate is scaled to have equal standard deviation (using the same scaling for both training and testing data).
For na\"ive Bayes, the R implementation from package e1071 is used.
For metric predictors, a Gaussian distribution (given the target class) is assumed.
This approach is less than ideal, but tangential to the theoretical target and experimental study.
For SVM, the R implementation from package e1071 is used. 
The SVM kernel used is radial basis kernel.
The probability calibration of the scores is performed for binary problems by MLE fitting of a logistic distribution to the decision values, or for multi-class problems, by computing the a-posteriori class probabilities using quadratic optimisation.







%

\vskip 0.2in

\bibliography{Shared}

\begin{thebibliography}{30}
\providecommand{\natexlab}[1]{#1}
\providecommand{\url}[1]{\texttt{#1}}
\expandafter\ifx\csname urlstyle\endcsname\relax
  \providecommand{\doi}[1]{doi: #1}\else
  \providecommand{\doi}{doi: \begingroup \urlstyle{rm}\Url}\fi

\bibitem[Adams et~al.(2010)Adams, Tasoulis, Anagnostopoulos, and
  Hand]{Adams2010}
Niall~M. Adams, Dimitris~K. Tasoulis, Christoforos Anagnostopoulos, and
  David~J. Hand.
\newblock Temporally-adaptive linear classification for handling population
  drift in credit scoring.
\newblock In \emph{In Proceedings of the 19th International Conference on
  Computational Statistics}, pages 167--176, 2010.

\bibitem[Anagnostopoulos et~al.(2012)Anagnostopoulos, Tasoulis, Adams, and
  Hand]{Anagnostopoulos2012}
Christoforos Anagnostopoulos, Dimitrios Tasoulis, Niall~M. Adams, and David~J.
  Hand.
\newblock Online linear and quadratic discriminant analysis with adaptive
  forgetting for streaming classification.
\newblock \emph{Statistical Analysis and Data Mining}, 5:\penalty0 139--166,
  2012.

\bibitem[Bache and Lichman(2013)]{Bache2013}
Kevin Bache and Moshe Lichman.
\newblock {UCI} machine learning repository, 2013.
\newblock URL \url{http://archive.ics.uci.edu/ml}.

\bibitem[Bishop(2007)]{Bishop2007}
Christopher Bishop.
\newblock \emph{Pattern Recognition and Machine Learning}.
\newblock Springer, 2007.

\bibitem[Breiman(2001)]{Breiman2001}
Leo Breiman.
\newblock Random forests.
\newblock \emph{Machine Learning}, 45(1):\penalty0 5--32, 2001.

\bibitem[Cortes and Vapnik(1995)]{Vapnik1995}
Corinna Cortes and Vladimir Vapnik.
\newblock Support-vector networks.
\newblock \emph{Machine Learning}, 20(3):\penalty0 273--297, 1995.

\bibitem[Dasgupta(2011)]{Dasgupta2011}
Sanjoy Dasgupta.
\newblock Two faces of active learning.
\newblock \emph{Theoretical Computer Science}, 412(19):\penalty0 1767--1781,
  2011.

\bibitem[Dasgupta and Hsu(2008)]{Dasgupta2008}
Sanjoy Dasgupta and Daniel Hsu.
\newblock Hierarchical sampling for active learning.
\newblock In \emph{Proceedings of the 25th International Conference of Machine
  Learning}, pages 208--215, 2008.

\bibitem[Duda et~al.(2001)Duda, Hart, and Stork]{Duda2001}
Richard Duda, Peter Hart, and David Stork.
\newblock \emph{Pattern Classification}.
\newblock Wiley, 2001.

\bibitem[Evans et~al.(2013)Evans, Adams, and Anagnostopoulos]{Evans2013}
Lewis P.~G. Evans, Niall~M. Adams, and Christoforos Anagnostopoulos.
\newblock When does active learning work?
\newblock In \emph{Intelligent Data Analysis Conference Proceedings, Lecture
  Notes in Computer Science Series by Springer-Verlag}, pages 174--185, 2013.

\bibitem[Geary(1954)]{Geary1954}
Robert~C. Geary.
\newblock The contiguity ratio and statistical mapping.
\newblock \emph{The Incorporated Statistician}, 5(3):\penalty0
  115--127,129--146, 1954.

\bibitem[Gu et~al.(2001)Gu, Hu, and Liu]{Gu2001}
Baohua Gu, Feifang Hu, and Huan Liu.
\newblock Modelling classification performance for large data sets.
\newblock \emph{Advances in Web-Age Information Management, Lecture Notes in
  Computer Science}, 2118:\penalty0 317--328, 2001.

\bibitem[Guyon et~al.(2011)Guyon, Cawley, Dror, and Lemaire]{Guyon2011}
Isabelle Guyon, Gavin Cawley, Gideon Dror, and Vincent Lemaire.
\newblock Results of the active learning challenge.
\newblock \emph{Journal of Machine Learning Research}, 16:\penalty0 19--45,
  2011.

\bibitem[Hand(1997)]{Hand1997}
David Hand.
\newblock \emph{Construction and Assessment of Classification Rules}.
\newblock Wiley, 1997.

\bibitem[Hand and Yu(2001)]{Hand2001}
David Hand and Keming Yu.
\newblock Idiot's {B}ayes: Not so stupid after all?
\newblock \emph{International Statistical Review}, 69(3):\penalty0 385--398,
  2001.

\bibitem[Hastie et~al.(2009)Hastie, Tibshirani, and Friedman]{Tibshirani2009}
Trevor Hastie, Robert Tibshirani, and Jerome Friedman.
\newblock \emph{The Elements of Statistical Learning}.
\newblock Springer, 2nd edition, 2009.

\bibitem[Hoi et~al.(2006)Hoi, Jin, Zhu, and Lyu]{Hoi2006}
Steven C.~H. Hoi, Rong Jin, Jianke Zhu, and Michael~R. Lyu.
\newblock Batch mode active learning and its application to medical image
  classification.
\newblock In \emph{Proceedings of the 23rd International Conference on Machine
  Learning}, pages 417--424, 2006.

\bibitem[Kadie(1995)]{Kadie1995}
Carl~Myers Kadie.
\newblock Seer: Maximum likelihood regression for learning-speed curves.
\newblock Submitted in partial fulfillment of the requirements for the degree
  of Doctor of Philosophy in Computer Science in the Graduate College of the
  University of Illinois at Urbana-Champaign, 1995.

\bibitem[Kumar et~al.(2010)Kumar, Ghani, Shak, Carbonell, and
  Rudnicky]{Kumar2010}
Mohit Kumar, Rayid Ghani, Mohak Shak, Jaime Carbonell, and Alexander Rudnicky.
\newblock Empirical comparison of active learning strategies for handling
  temporal drift.
\newblock \emph{ACM Transactions on Embedded Computing Systems}, 9(4):\penalty0
  161--168, 2010.

\bibitem[Mitchell(1997)]{Mitchell1997}
Tom Mitchell.
\newblock \emph{Machine Learning}.
\newblock McGraw-Hill Higher Education, 1997.

\bibitem[Moran(1950)]{Moran1950}
Patrick A.~P. Moran.
\newblock Notes on continuous stochastic phenomena.
\newblock \emph{Biometrika}, 37:\penalty0 17--23, 1950.

\bibitem[Olsson(2009)]{Olsson2009}
Fredrik Olsson.
\newblock A literature survey of active machine learning in the context of
  natural language processing.
\newblock Technical Report ISSN: 1100-3154, Swedish Institute of Computer
  Science, 2009.

\bibitem[Perlich et~al.(2003)Perlich, Provost, and Simonoff]{Provost2003}
Claudia Perlich, Foster Provost, and Jeffrey~S. Simonoff.
\newblock Tree induction vs. logistic regression: A learning-curve analysis.
\newblock \emph{Journal of Machine Learning Research}, 4:\penalty0 211--255,
  2003.

\bibitem[Provost and Attenberg(2010)]{Provost2010}
Foster Provost and Josh Attenberg.
\newblock Inactive learning difficulties employing active learning in practice.
\newblock \emph{ACM SIGKDD}, 12:\penalty0 36--41, 2010.

\bibitem[Ripley(1996)]{Ripley1996}
Brian~D. Ripley.
\newblock \emph{Pattern Recognition and Neural Networks}.
\newblock Cambridge University Press, 1996.

\bibitem[Roy and Mccallum(2001)]{Roy2001}
Nicholas Roy and Andrew Mccallum.
\newblock Toward optimal active learning through sampling estimation of error
  reduction.
\newblock In \emph{Proceedings of the 18th International Conference on Machine
  Learning}, pages 441--448, 2001.

\bibitem[Schein and Ungar(2007)]{Schein2007}
Andrew Schein and Lyle Ungar.
\newblock Active learning for logistic regression: an evaluation.
\newblock \emph{Machine Learning}, 68(3):\penalty0 235--265, 2007.

\bibitem[Settles(2009)]{Settles2009}
Burr Settles.
\newblock Active learning literature survey.
\newblock Computer Sciences Technical Report 1648, University of
  Wisconsin--Madison, 2009.

\bibitem[Seung et~al.(1992)Seung, Opper, and Sompolinsky]{Seung1992}
H.~Sebastian Seung, Manfred Opper, and Haim Sompolinsky.
\newblock Query by committee.
\newblock In \emph{Proceedings of the 5th annual workshop on Computational
  Learning Theory}, pages 287--294, 1992.

\bibitem[Thrun and M{\"o}ller(1992)]{Thrun1992}
Sebastian Thrun and Knut M{\"o}ller.
\newblock Active exploration in dynamic environments.
\newblock In \emph{Proceeding of the 25th Conference of Advances in Neural
  Information Processing Systems}, volume~4, pages 531--538, 1992.

\end{thebibliography}

\end{document}